\documentclass[10pt,journal,compsoc,x11names]{IEEEtran}

\usepackage[switch]{lineno}
\usepackage{amsmath}
\usepackage{xspace}
\usepackage{enumitem}
\usepackage{amssymb}
\usepackage{booktabs}       
\usepackage[hyphens]{url}
\usepackage{ragged2e}
\usepackage{color}
\usepackage{colortbl}
\usepackage{tablefootnote}
\usepackage{pifont}
\usepackage{makecell}
\usepackage{varwidth}
\usepackage{pdflscape}
\usepackage[most]{tcolorbox}
\usepackage{framed}
\usepackage{mdframed}
\usepackage{caption}
\usepackage{longtable}
\usepackage{float}

\usepackage[utf8]{inputenc} 
\usepackage[T1]{fontenc}    
\usepackage{hyperref}       
\usepackage{url}            

\usepackage{amsfonts}       
\usepackage{nicefrac}       
\usepackage{microtype}      
\usepackage{xcolor}         

\usepackage{algorithmic}
\usepackage{algorithm}
\usepackage{multirow}
\usepackage{bm}
\usepackage{subcaption}
\usepackage{graphicx}
\usepackage{wrapfig}
\usepackage{hyperref}       
\usepackage{url}            

\newcommand{\paratitle}[1]{\vspace{1.5ex}\noindent\textbf{#1}}
\newcommand{\paraitem}[1]{$\bullet$ \textbf{#1}}

\definecolor{gold}{RGB}{205,133,63}
\definecolor{fGreen}{RGB}{34,139,34}
\definecolor{tOrange}{RGB}{255,215,0}
\definecolor{tBlue}{RGB}{135,206,250}
\definecolor{tPink}{RGB}{255,204,204}
\definecolor{tGreen}{RGB}{205,230,199}
\definecolor{tGold}{RGB}{255,215,0}

\usepackage{cite}
\usepackage[numbers,sort&compress]{natbib}

\ifCLASSINFOpdf
\else
\fi

\begin{document}
\title{Examining User-Friendly and Open-Sourced Large GPT Models: A Survey on Language, Multimodal, and Scientific GPT Models}

\author{Kaiyuan Gao, Sunan He, Zhenyu He, Jiacheng Lin, QiZhi Pei, Jie Shao, Wei Zhang
\IEEEcompsocitemizethanks{
\IEEEcompsocthanksitem All authors contribute equally to this work, and the author list is arranged alphabetically.
\IEEEcompsocthanksitem Kaiyuan Gao is with Huazhong University of Science and Technology (im\_kai@hust.edu.cn).
\IEEEcompsocthanksitem Sunan He is with The Hong Kong University of Science and Technology (sunan.he@connect.ust.hk).
\IEEEcompsocthanksitem Zhenyu He is with Peking University (hezhenyu@stu.pku.edu.cn).
\IEEEcompsocthanksitem Jiacheng Lin is with University of Illinois Urbana-Champaign (jl254@illinois.edu).
\IEEEcompsocthanksitem Qizhi Pei is with Renmin University of China (qizhipei@ruc.edu.cn). 
\IEEEcompsocthanksitem Jie Shao is with Nanjing University (shaoj@lamda.nju.edu.cn). 
\IEEEcompsocthanksitem Wei Zhang is with University of Science and Technology of China (weizhang\_cs@mail.ustc.edu.cn). 
\IEEEcompsocthanksitem More models are continuously updated in the github: \url{https://github.com/GPT-Alternatives/gpt_alternatives}, such as multiple LLaMA2-based open-source models.  
}%
}

\markboth{}%
{Shell \MakeLowercase{\textit{et al.}}: Bare Advanced Demo of IEEEtran.cls for IEEE Computer Society Journals}

\IEEEtitleabstractindextext{%
\begin{abstract}
\justifying
Generative pre-trained transformer (GPT) models have revolutionized the field of natural language processing (NLP) with remarkable performance in various tasks and also extend their power to multimodal domains. Despite their success, large GPT models like GPT-4 face inherent limitations such as considerable size, high computational requirements, complex deployment processes, and closed development loops. These constraints restrict their widespread adoption and raise concerns regarding their responsible development and usage. The need for user-friendly, relatively small, and open-sourced alternative GPT models arises from the desire to overcome these limitations while retaining high performance. In this survey paper, we provide an examination of alternative open-sourced models of large GPTs, focusing on user-friendly and relatively small models that facilitate easier deployment and accessibility. The main contents of this paper are divided into the following itemized items:
(1) Investigate the architecture, design principles, and trade-offs of user-friendly and relatively small alternative GPT models, focusing on their ability to overcome the challenges posed by large GPT models.
(2) Present the data collection and analyze the pre-training data source, data quality, quantity, diversity, finetuning data including instruction data, alignment data, and also the domain-specific data for domain-specific models. 
(3) Survey the techniques for efficient deployment and fine-tuning of these GPT models.
(4) Introduce ongoing open-source projects and initiatives for user-friendly GPT model reproduction and deployment. 
(5) Provide a thorough analysis of benchmark evaluations and offer human evaluations of these relatively small GPT models to give some human-liked recommendations in real usage. 
(6) Explore the extension of GPT models to multimodal settings, focusing on models that integrate NLP with computer vision, and also place special focus on user-friendly scientific GPT models and biomedical domains.
Through this extensive survey, we aim to equip researchers, practitioners, and enthusiasts with a thorough understanding of user-friendly and relatively small open-sourced models of large GPTs, their current state, challenges, and future research directions, inspiring the development of more efficient, accessible, and versatile GPT models that cater to the broader scientific community and advance the field of general artificial intelligence.
The source contents are continuously updating in \url{https://github.com/GPT-Alternatives/gpt_alternatives}.
\end{abstract}

}

\maketitle

\IEEEdisplaynontitleabstractindextext

\IEEEpeerreviewmaketitle

\section{Introduction}

The advent of generative pre-trained transformer (GPT) models has brought about a significant transformation in the field of natural language processing (NLP). These models, based on the transformer \cite{DBLP:conf/nips/VaswaniSPUJGKP17} architecture, demonstrate exceptional capabilities in various NLP tasks \cite{brown2020language,openai2023gpt}. The continuous development of GPT models has led to increasingly larger and more sophisticated versions, with large GPT models like GPT-4~\cite{openai2023gpt} gaining significant attention for their unparalleled performance. However, despite their impressive capabilities, large GPT models have inherent limitations that restrict their widespread adoption, usability, and fine-tuning.

The considerable size of these models leads to high computational requirements, extensive memory usage, and complex deployment processes. For example, the latest model GPT-4~\cite{openai2023gpt} was trained using an unprecedented scale of computing and data. These constraints not only hinder the accessibility of large GPT models for researchers and practitioners with limited resources but also raise concerns regarding their energy consumption and environmental impact. In addition to these resource-related challenges, large GPT models often face issues related to their training data, including the potential to generate biased or inappropriate content, the reinforcement of stereotypes, and the lack of transparency in data collection and preprocessing. Addressing these limitations is crucial to ensure the responsible development and deployment of GPT models in various applications and domains.

The need for alternative GPT models arises from the desire to overcome the aforementioned limitations while retaining the high performance demonstrated by the original large GPT models. In this survey paper, we conduct an in-depth examination of user-friendly, relatively small, and open-sourced models of large GPTs, focusing on their architecture, efficiency, deployment strategies, and fine-tuning methods. By exploring these alternative models, we aim to provide insights into their potential to address the challenges posed by the original large GPT models and facilitate the development of more efficient, accessible, and responsible NLP technologies.

We begin with a thorough overview of these open-sourced GPT models, discussing their unique features, design principles, and the trade-offs encountered during their development. We emphasize the importance of efficiency in these models by exploring techniques to minimize model size, memory usage, and computational demands without sacrificing performance. This analysis allows us to identify promising approaches that can help mitigate the limitations of large GPT models in terms of resource requirements and environmental impact.

In addition to resource efficiency, we investigate various aspects of data that play a critical role in the development of these GPT models. We discuss the pre-training data sources, which serve as the foundation for training these models, followed by an exploration of data quality, quantity, and diversity, which are essential factors influencing model performance \cite{brown2020language, chowdhery2022palm, DBLP:journals/corr/abs-2112-11446, scao2022bloom}. Furthermore, we examine fine-tuning data, including instruction data and alignment data, which are crucial for refining models and ensuring their adaptability to specific tasks \cite{zhou2023lima, ouyang2022training, selfinstruct, DBLP:conf/iclr/WeiBZGYLDDL22}. Lastly, we explore domain-specific data, which is vital for training models tailored to address challenges and applications in specialized fields.

Subsequently, we investigate the methods and approaches for deploying and fine-tuning alternative GPT models, addressing the challenges associated with deployment, such as hardware limitations \cite{yao2022zeroquant}. We also discuss strategies for fine-tuning, including zero-shot and few-shot learning, which enable these models to be customized for specific tasks and domains while reducing the risk of generating biased or inappropriate content \cite{houlsby2019parameter, hu2021lora, dettmers2023qlora, lester2021power, liu2021p}.
Besides, we also delve into different open-sourced efforts and tool development. We examine the open-sourced initiatives, libraries \cite{wolf2020transformers, rasley2020deepspeed, peft}, and platforms \cite{open-llm-leaderboard, dubois2023alpacafarm} that facilitate the development, sharing, and collaboration of GPT models within the research community, fostering innovation and the development of more efficient, accessible, and versatile models.

Importantly, we place significant emphasis on both benchmark evaluation and human evaluation of these open-sourced GPT alternatives, ensuring a thoughtful assessment of their performance. Specifically, we test these models across a range of well-known benchmark datasets from diverse domains~\cite{Clark2018ThinkYH, clark2019boolq, zellers2019hellaswag, sakaguchi2021winogrande, Bisk2020}, providing a robust comparison of their capabilities. In addition, we conduct human evaluations by collecting an array of diverse, meaningful, and representative instructions, which enables us to gather valuable real-world feedback. The thorough results analysis offers valuable insights into the strengths and weaknesses of these models.

We also explore the emerging fields of multimodal \cite{zhu2023minigpt, liu2023visual, zhang2023transfer} and scientific GPT models \cite{DBLP:journals/corr/abs-2211-09085, DBLP:journals/bib/LuoSXQZPL22}, highlighting their potential applications and performance in specialized domains. 
Finally, we summarize the whole survey, and reflect on the broader implications of alternative GPT models for the field of artificial intelligence and society at large, with a discussion about the existing challenges and limitations of the GPT models, and highlighting the potential future directions for developing more efficient, accessible, reliable, and versatile GPT models.

\begin{figure*}[tbp]
    \centering
    \includegraphics[width=0.98\linewidth]{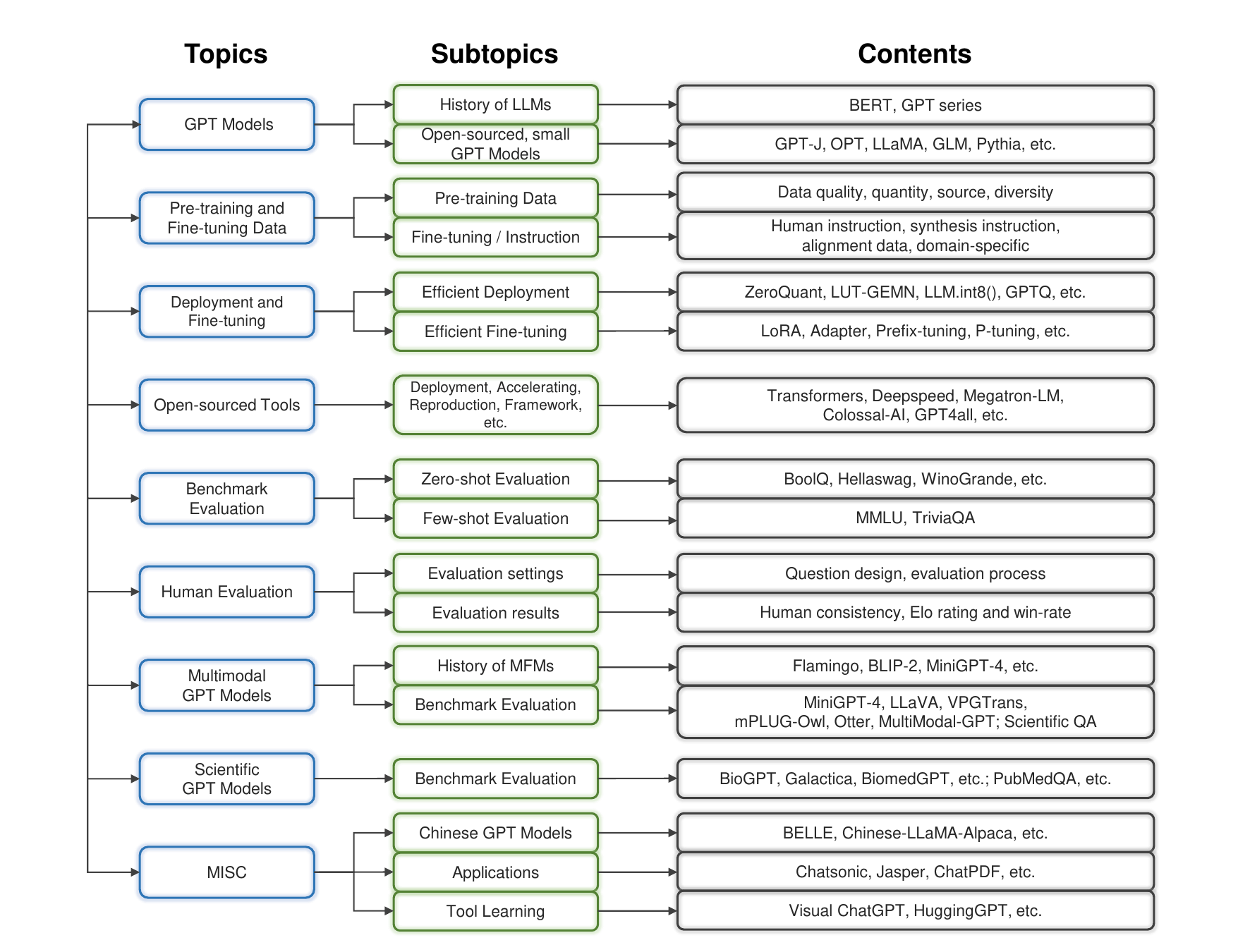}
    \caption{Overview of the paper's general structure.}
    \label{fig:overview}
    \vspace{-0.5cm}
\end{figure*}

The overview and the general structure of this paper is illustrated in Figure~\ref{fig:overview}.

\section{GPT and GPT-like Models}
There are three requirements for an alternative model that we investigate: (1) It is pre-trained on a large amount of data and the model size is relatively small which enables user-friendly utilization (near $10$B parameters). (2) It ensures the basic requirement that can generate (mostly text-based) content. (3) The model checkpoint is accessible so practitioners can use it for further study and downstream applications. 


\subsection{History of Large Language Foundation Models}
\begin{figure*}[tbp]
    \centering
    \includegraphics[width=1.0\linewidth]{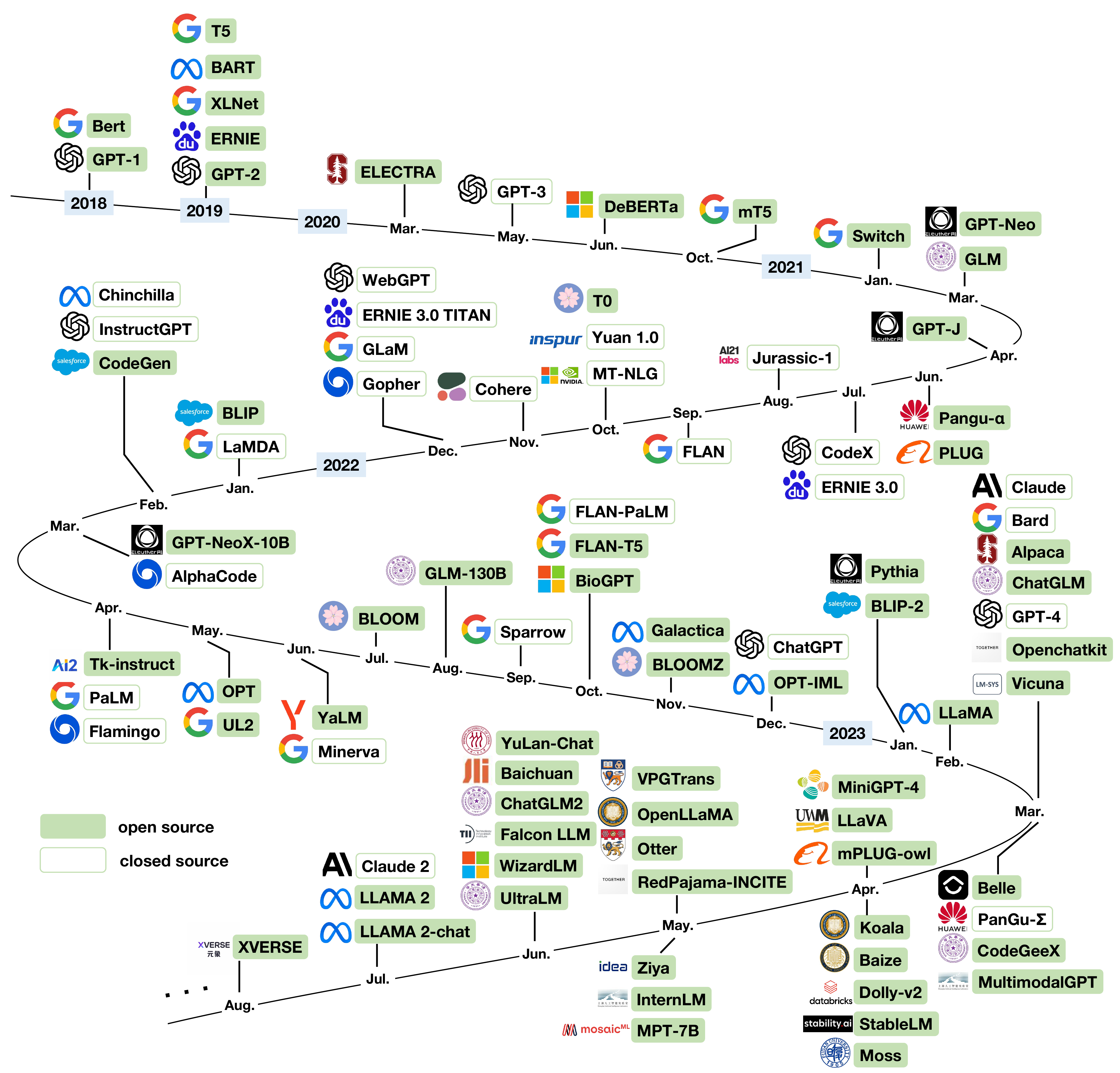}
    \caption{A chronological overview of large language models (LLMs), multimodal and scientific models in recent years. LLMs with publicly available model checkpoints are highlighted in green. }
    \label{fig:evolution}
    \vspace{-0.5cm}
\end{figure*}

With the emergence of the Transformer~\cite{vaswani2017attention} model, there has been a significant change in the field of natural language processing. The Transformer model is superior not only in performance (e.g., for machine translation~\cite{vaswani2017attention}), but also more parallelizable than RNN~\cite{elman1990finding} models like LSTM~\cite{hochreiter1997long} and GRU~\cite{cho2014learning}. GPT-1~\cite{radford2018improving} and Bert~\cite{devlin2018bert} were among the first attempts to utilize the Transformer model for unsupervised pretraining. By pretraining on large amount of unlabeled data and finetuning on downstream data, they became state-of-the-art models in NLP leaderboards. The different pretraining objectives for Bert and GPT-1, i.e., masked language modeling and next token prediction, contribute to their expertise. When compared at a similar model size by finetuning, BERT has been observed to perform better on language understanding tasks than GPT, while GPT is better suited for language generation tasks. With the evolution of GPT models from GPT-1~\cite{radford2018improving} and GPT-2~\cite{radford2019language} to GPT-3~\cite{brown2020language}, two outstanding traits of autoregressive language models have been observed: (1) By increasing the model size and pretraining data size, performance can be boosted. (2) GPT-3 shows extraordinary few-shot and zero-shot performance by performing in-context learning and prompting. Since the success of GPT-3, there have been more and more large language models developed. One milestone afterward in the development of large language models has been the emergence of ChatGPT\footnote{\url{https://chat.openai.com/}}, a conversational model finetuned with reinforcement learning from human feedback (RLHF) and built on GPT-3.5, an upgraded version of GPT-3. Through the alignment of human preferences, ChatGPT exhibits remarkable conversational proficiency and gains rapid recognition among the general public. Following the remarkable achievements of ChatGPT, the community has recently witnessed the release of GPT-4~\cite{openai2023gpt}. GPT-4 has demonstrated improved conversational proficiency, enhanced contextual understanding, and more accurate and coherent responses. By employing cutting-edge techniques and building on the successes of ChatGPT and GPT-3.5, GPT-4 has made significant strides in multiple domains (e.g., multimodal), which marks another milestone in the evolution of large language models, showcasing the potential for even more groundbreaking innovations and applications in the near future. Along the process, huge efforts from the whole AI community has been paid to develop GPT-like large language models, either aims to build stronger models from big company (e.g., Bard\footnote{\url{https://bard.google.com/}}) or create open-sourced replacements (e.g., LLaMA~\cite{touvron2023llama}) to boost the development of foundation models. An overview history of these large language models (LLMs) in recent years is shown in Figure~\ref{fig:evolution}.




\subsection{Open Language Foundation Models}
\begin{table*}
\centering
\caption{Statistical overview of open large language models in recent years, categorized by base models.}
\label{tab2:stat_models}
\scalebox{0.8}{
\begin{tabular}{l|p{0.3 \columnwidth}|p{0.3 \columnwidth}|p{0.15 \columnwidth}|p{0.7 \columnwidth}|p{0.3 \columnwidth}}  
\toprule
Model & \#Param & Backbone & Release Date & Training Data Source & Training Data Size \\
\midrule
T5~\cite{raffel2020exploring} (enc-dec) & 60M, 220M, 770M, 3B, 11B & Base Model & 2019-10 & C4~\cite{raffel2020exploring} & 1T tokens \\
\midrule
mT5~\cite{xue2020mt5} (enc-dec)  & 300M, 580M, 1.2B, 3.7B, 13B & Base Model & 2020-10 & mC4 & 1T tokens \\
\midrule
GPT-Neo~\cite{gpt-neo} & 125M, 350M, 1.3B, 2.7B & Base Model & 2021-03 & the Pile~\cite{gao2020pile} & 825GB \\
GPT-NeoX~\cite{black2022gpt} & 20B & Base Model & 2022-02 & the Pile~\cite{gao2020pile} &  825GB \\
\midrule
GPT-J~\cite{gpt-j} & 6B & Base Model & 2021-06 & the Pile~\cite{gao2020pile} & 825GB \\
\midrule
OPT~\cite{zhang2022opt} & 125M, 1.3B, 2.7B, 6.7B, 13B, 30B, 66B, 175B & Base Model & 2022-05 & the Pile~\cite{gao2020pile} and d PushShift.io Reddit~\cite{baumgartner2020pushshift} & 180B tokens \\
\midrule
BLOOM~\cite{scao2022bloom} & 560M, 1.1B, 1B7, 3B, 7.1B, 176B & Base Model & 2022-07 & ROOTS corpus~\cite{laurenccon2022bigscience} & 366B tokens \\
BLOOMZ & 560M, 1.1B, 1B7, 3B, 7.1B, 176B & BLOOM & 2022-11 & xP3(extended from P3~\cite{sanh2021multitask}) & - \\
\midrule
GLM~\cite{du2022glm} & 110M, 335M, 410M, 515M, 2B, 10B, 130B & Base Model & 2021-03 & BooksCorpus~\cite{zhu2015aligning} and
English Wikipedia & - \\
GLM-130B~\cite{zeng2023glm-130b} & 130B & Base Model &  2022-08 & - & - \\
ChatGLM ~\cite{du2022glm, zeng2023glm-130b} & 6B & GLM & 2023-03 & - & - \\
ChatGLM2 ~\cite{du2022glm, zeng2023glm-130b} & 6B & GLM & 2023-06 & - & - \\
\midrule
LLaMA~\cite{touvron2023llama} & 7B, 13B, 33B, 65B & Base Model & 2023-02 & English CommonCrawl, C4~\cite{raffel2020exploring}, Github, Wikipedia, Gutenberg Books3 and Stack Exchange & 1.4T tokens \\
OpenLLaMA~\cite{openlm2023openllama} & 3B, 7B & Replicate of LLaMA & 2023-05 & & \\
Alpaca~\cite{alpaca} & 7B & LLaMA & 2023-03 & data generated from text-davinci-003 & 52K \\
Vicuna~\cite{vicuna} & 7B, 13B & LLaMA & 2023-03 & user-shared conversations from ShareGPT & 70K \\
StableVicuna~\cite{stablelm} & 13B & LLaMA \& Vicuna & 2023-04 & - & - \\
BAIZE~\cite{xu2023baize} & 7B, 13B, 30B & LLaMA & 2023-04 &  dialogs from Quora, StackOverFlow and MedQuAD questions & 54K/57K/47K \\
Koala~\cite{koala_blogpost_2023} & 13B & LLaMA & 2023-04 & A gather of ShareGPT\footnote{\url{https://sharegpt.com}}, HC3~\cite{guo2023close}, OIG\footnote{\url{https://laion.ai/blog/oig-dataset}}, Stanford Alpaca~\cite{alpaca}, Anthropic HH\footnote{\url{https://huggingface.co/datasets/Anthropic/hh-rlhf}}, OpenAI WebGPT~\cite{nakano2021webgpt}, and OpenAI Summarization~\cite{stienon2020learning} & - \\
WizardLM~\cite{xu2023wizardlm} & 7B, 13B, 30B & LLaMA & 2023-06 & evolved instructions (from ShareGPT)/evolved instructions (from Alpaca~\cite{alpaca} data) & 250k/70k \\
UltraLM~\cite{ding2023ultralm} & 13B & LLaMA & 2023-06 & UltraChat~\cite{UltraChat} & - \\
YuLan-Chat~\cite{YuLan-Chat} & 13B, 65B & LLaMA & 2023-06 & - & - \\
\midrule
Pythia~\cite{biderman2023pythia} & 70M, 160M, 410M, 1B, 1.4B, 2.8B, 6.9B, 12B & Base Model & 2023-01 & the Pile~\cite{gao2020pile}/the Pile with deduplication applied & 299.9B tokens/207B tokens \\
Dolly-v2~\cite{dolly} & 12B & Pythia & 2023-04 & instruction/response finetuning records & \textasciitilde 15k \\
Openchatkit~\cite{openchatkit} & 7B & Pythia & 2023-03 & the OIG\footnote{\url{https://laion.ai/blog/oig-dataset/}} dataset & \\
BELLE-7B~\cite{belle} & 7B & Pythia & 2023-03 & a Chinese Dataset\footnote{\url{https://github.com/LianjiaTech/BELLE/tree/main/data/1.5M}} & 1.5M \\
\midrule
StableLM-Alpha~\cite{stablelm} & 3B, 7B & Base Model & 2023-04 & dataset that build on the Pile~\cite{gao2020pile} & 1.5T tokens \\
StableLM-Tuned-Alpha~\cite{stablelm} & 7B & StableLM & 2023-04 & Stanford's Alpaca, Nomic-AI's gpt4all, RyokoAI's ShareGPT-52K datasets \cite{sharegpt52K}, Databricks labs' Dolly, and Anthropic's HH & - \\
\midrule
RWKV~\cite{peng2023rwkv} & 169M, 430M, 1.5B, 3B, 7B, 14B & Base Model & - & the Pile~\cite{gao2020pile} & 825GB \\
ChatRWKV~\cite{chatrwkv} & 7B, 14B & RWKV & 2022-12 & - & -  \\
\midrule
moss-moon-003-base~\cite{moss} & 16B & base model & 2023-04 & - & 700B tokens \\
moss-moon-003-sft~\cite{moss} & 16B & moss-moon-003-base  & 2023-04 & multi-round conversational data & 1.1 million \\
\midrule
RedPajama-INCITE~\cite{RedPajama-INCITE} & 3B, 7B & Base Model & 2023-05 & RedPajama-Data~\cite{redpajama-data} & 1.2T tokens \\
\midrule
MPT-7B~\cite{mpt7b}  & 7B & Base Model & 2023-05 & - & 1T tokens \\
MPT-7B-Chat & 7B & MPT-7B & 2023-05 &  ShareGPT-Vicuna, HC3, Alpaca, Helpful and Harmless, and Evol-Instruct datasets & - \\
\midrule
Falcon LLM~\cite{falcon} & 7B, 40B & Base Model & 2023-06 & - & 1T tokens \\
\midrule
InternLM~\cite{2023internlm} & 7B & Base Model & 2023-06 & - & trillions of tokens \\
InternLM Chat~\cite{2023internlm} & 7B & InternLM & 2023-06 & - & - \\
\midrule
Baichuan~\cite{baichuan} & 7B & Base Model & 2023-06 & - & 1.2T tokens \\
\midrule
LLAMA 2~\cite{touvron2023llama_2} & 7B, 13B, 70B & Base Model & 2023-07 & a mix of data from publicly available sources & 2T tokens \\
LLAMA 2-CHAT~\cite{touvron2023llama_2} & 7B, 13B, 70B & LLAMA 2 & 2023-07 & publicly available instruction tuning
data and vendor-based annotation data& 27,540 instruction tuning data, 2,919,326 human preference data \\
\midrule
Qwen~\cite{2023qwen} & 7B & Base Model & 2023-08 & - & 2.2T tokens \\
Qwen-Chat~\cite{2023qwen} & 7B & Qwen & 2023-08 & - & - \\
\midrule
XVERSE~\cite{XVERSE-13B} & 13B & Base Model & 2023-08 & - & 1.4T tokens \\

\bottomrule
\end{tabular} 
}
\end{table*}

In this section, we introduce the efforts of open-sourced language models that developed by the whole community. For these language foundation models, we mainly focus on the following aspects:
\begin{enumerate}
    \item \textbf{Model Structure}: 
    Transformer~\cite{vaswani2017attention} architecture has become the universal architecture for large language models. There are three main adaptions of the Transformer architecture, i.e., encoder-only, decoder-only, and encoder-decoder. As the name indicates, encoder-only Transformer only utilizes the encoder part, such as Bert~\cite{devlin2018bert}, ERNIE~\cite{sun2019ernie}, ELECTRA~\cite{clark2020electra} and so on. Decoder-only Transformer only utilizes the decoder part, such as GPT series~\cite{radford2018improving,radford2019language,brown2020language}, OPT~\cite{zhang2022opt}, BLOOM~\cite{scao2022bloom}, PaLM~\cite{chowdhery2022palm}, LLaMA~\cite{zhang2023llama} and so on. Encoder-decoder Transformer utilizes the whole Transformer architecture, such as T5~\cite{raffel2020exploring}, Bart~\cite{lewis2019bart}, GLM~\cite{du2021glm}, and so on. Notably, the decoder-based GPT-like models are the main focus of this survey paper.
    
    \item \textbf{Pretraining Dataset}: The performance and generalization capabilities of models are significantly influenced by the quality and size of the pretraining data. The focus is mainly on the public datasets. Two commonly employed sources of datasets are web crawling and books/literature. Taking LLaMA~\cite{touvron2023llama} as an example, the pretraining dataset is a mixture of many publicly available sources, including English CommonCrawl, C4~\cite{raffel2020exploring}, Github, Wikipedia, Gutenberg and Books3, and Stack Exchange. 
    
    \item \textbf{Pretraining Tasks}: Language modeling, aka., next token prediction, is the dominant pretraining task in large language models. It's firstly observed from GPT-3~\cite{brown2020language} that scaling model sizes and pretraining data can vastly increase the model's few-shot/zero-shot ability. Other representative models using this pretraining task include PaLM~\cite{chowdhery2022palm}, LLaMA~\cite{touvron2023llama}, and so on. Earlier works, e.g. Bert~\cite{devlin2018bert}, RoBERTa~\cite{liu2019roberta} adopted masked language modeling objectives. Although these models are good at natural language understanding, they are inferior in language generation and few-shot/zero-shot ability. Other works such as T5~\cite{raffel2020exploring} and GLM~\cite{du2021glm} use denoising pretraining objectives.  
\end{enumerate}

Next, we will introduce several language models within the above aspects that are open-sourced. A history summary of these models is in Figure~\ref{fig:evolution}.

\textbf{T5}~\cite{raffel2020exploring} is an encoder-decoder transformer model pre-trained with 1T tokens on C4~\cite{raffel2020exploring}. The pretraining objective is a denoising objective, i.e. to mask consecutive spans of tokens and only predict dropped-out
tokens. The released model checkpoints include 60M, 220M, 770M, 3B, and 11B. Flan-T5~\cite{chung2022scaling} is finetuned on chain-of-thought data based on T5, which the performance on downstream tasks is much better compared with T5.

\textbf{mT5}~\cite{xue2020mt5} is a multilingual encoder-decoder transformer model pre-trained with 1T tokens covering 101 languages. The pretraining objective is the same as T5. The released pre-trained model checkpoints include 300M, 580M, 1.2B, 3.7B, and 13B.

\textbf{GPT-J}~\cite{gpt-j} is a 6-billion parameter open-source English autoregressive language model (GPT-like) trained on the Pile~\cite{gao2020pile}. It's a decoder-only model trained with the next token prediction objective.

\textbf{GPT-Neo}~\cite{gpt-neo} is an implementation of GPT3-like models. It's a decoder-only model trained with the next token prediction objective and trained on the Pile~\cite{gao2020pile} dataset. The released pre-trained model checkpoints include 125M, 350M, 1.3B, and 2.7B.
\textbf{GPT-NeoX-20B}~\cite{black2022gpt} is a larger extension version of GPT-Neo, the released is a 20-billion parameter model. 

\textbf{OPT}~\cite{zhang2022opt} is a suite of decoder-only pre-trained transformers. The training objective is also the next token prediction. The released model checkpoints include 125M, 1.3B, 2.7B, 6.7B, 13B, 30B, 66B, and 175B.

\textbf{Bloom}~\cite{scao2022bloom} is an open-access multilingual language model. It's a decoder-only model trained with the next token prediction objective. The released model checkpoints include 560M, 1B1, 1B7, 3B, 7B1, and 176B. \textbf{BloomZ} is a multitask prompted finetuning model based on Bloom.

\textbf{GLM}~\cite{du2022glm, zeng2023glm-130b} is a pre-trained encoder-decoder model with an autoregressive blank infilling objective. The released pre-trained model checkpoints include 110M, 335M, 410M, 515M, 2B, 10B, and 130B. \textbf{ChatGLM-6B} and \textbf{ChatGLM2-6B} are two open-source bilingual(English and Chinese) chat models finetuned on GLM. 

\textbf{LLaMA}~\cite{touvron2023llama} is an open-source autoregressive language model similar to GPT-3 with moderate architecture modifications. The released pre-trained model checkpoints include 7B, 13B, 33B, and 65B. Since LLaMA is the first well-acknowledged open-sourced large language model with satisfied performance (comparable to GPT-3), there are many open-source subsequent (instruction) finetuned ChatGPT-like models based on LLaMA, such as \textbf{Alpaca-7B}~\cite{alpaca}, \textbf{Vicuna-7B and 13B}~\cite{vicuna}, ~\cite{du2022glm, zeng2023glm-130b}~\cite{stablelm}, \textbf{BAIZE-7B, 13B and 30B}~\cite{xu2023baize}, \textbf{Koala-13B}~\cite{koala_blogpost_2023}, \textbf{WizardLM}~\cite{xu2023wizardlm}, \textbf{UltraLM}~\cite{ding2023ultralm}, YuLan-Chat~\cite{YuLan-Chat}. Besides, \textbf{OpenLLaMA}~\cite{openlm2023openllama} is a reproduction of LLaMA which can be used for commercial purposes. Most recently, \textbf{LLAMA }2~\cite{touvron2023llama_2} is an updated version of LLaMA with more pretraining data and doubled context length. \textbf{LLAMA 2-Chat}~\cite{touvron2023llama_2} is a finetuned version of LLAMA 2 for dialogue. The released checkpoints of LLAMA 2 and LLAMA 2-CHAT include 7B, 13B, and 70B. LLAMA2 series can also be used for commercial purposes.

\textbf{Pythia}~\cite{biderman2023pythia} is also an autoregressive language model similar to GPT-3 with moderate architecture modifications. The released pre-trained model checkpoints include 70M, 160M, 410M, 1B, 1.4B, 2.8B, 6.9B, and 12B. There are many open-source subsequent (instruction) finetuned ChatGPT-like models based on Pythia, such as \textbf{Dolly-v2-12b~\cite{dolly}}, \textbf{Openchatkit-7B~\cite{openchatkit}}, \textbf{BELLE-7B~\cite{belle}}.

\textbf{StableLM-Alpha}~\cite{stablelm} is an ongoing series of GPT-like language models. The released pre-trained model checkpoints of StableLM-Alpha include 3B and 7B. \textbf{StableLM-Tuned-Alpha}~\cite{stablelm} is a series of finetuned models based on StableLM-Alpha for conversational purposes, the tuned datasets are in various formats. The released pre-trained model checkpoints of StableLM-Tuned-Alpha include 3B and 7B. 

\textbf{ChatRWKV}~\cite{chatrwkv} is an open-sourced ChatGPT-like model based on the RWKV~\cite{peng2023rwkv} language model, which is an RNN model rather than a Transformer model. The released model checkpoints include 7B and 14B.

\textbf{Moss}~\cite{moss}, similar to ChatGPT, is a conversational language model that can execute diverse natural language tasks, such as generating text, answering questions, summarizing text, generating code, and more, by following users' instructions. The released model checkpoints include 16B.

\textbf{RedPajama-INCITE}~\cite{RedPajama-INCITE} models are a series of models that aim to replicate the LLaMA recipe closely with fully open-sourced instruction-tuned and chat models based on them. The release model checkpoints include 3B and 7B.

\textbf{MPT-7B}~\cite{mpt7b} is open source and can be used for commercial purposes, and offers performance that is on par with LLaMA-7B. \textbf{MPT-7B-Instruct} and \textbf{MPT-7B-Chat} are finetuned models based on MPT-7B.

\textbf{Baichuan}~\cite{baichuan} is an open-sourced GPT language model trained on about 1.2T tokens. It's available for commercial purposes. The release checkpoints include 7B.

\textbf{Falcon LLM}~\cite{falcon} is an autoregressive decoder-only language model. It's open-sourced and allows for commercial use with no restrictions. The release model checkpoints include 7B and 40B.

\textbf{InternLM}~\cite{2023internlm} is an open-sourced foundation model pretrained on trillions of high-quality tokens. The release model checkpoints include 7B. \textbf{InternLM Chat} is a finetuned chat model based on InternLM. The release model checkpoints include 7B.

\textbf{Qwen}~\cite{2023qwen} is an open-sourced decoder-only model pretrained on over 2.2 trillion tokens. The released model checkpoints of Qwen include 7B. \textbf{Qwen-Chat} is a fine-tuned human-aligned version based on Qwen. The released model checkpoints of Qwen-Chat include 7B.

\textbf{XVERSE}~\cite{XVERSE-13B} is an open-sourced decider-only model pretrained on 1.4 trillion tokens. The released model checkpoints of XVERSE include 13B.
 

\subsection{Evaluation Models}
In this survey, a large and main contribution is that we explore the potential of various GPT-like open models by conducting a comprehensive evaluation across multiple dimensions. These dimensions include general language benchmarks, which are the main focus of these language models, and also scientific domain datasets and multimodal datasets that perform as additional evaluation of these multimodal foundation models and scientific GPT models. Another important one is human evaluation that we paid much effort. Therefore, the evaluated models are listed in Table \ref{tab2:eval_models} for language models and scientific language models, and Table \ref{tab2:eval_models_MM} for multi-modal models. All the models under evaluation were sourced from the Huggingface \cite{wolf2020transformers} or the original Github repositories. 

In total, we conduct evaluations on 32 open-sourced models with different model sizes around $10$B. More specifically, 24 models from Table~\ref{tab2:eval_models} on language and scientific GPT models, and 8 models from Table~\ref{tab2:eval_models_MM} on multimodal GPT models for benchmark evaluations. For human evaluation, we evaluate 16 of these models to have a relatively comprehensive study to give more convincing results. 

\begin{table*}[!ht]
    \centering
    \caption{Overview of our evaluated language and scientific GPT models.}
    \label{tab2:eval_models}
    \begin{tabular}{lcl}
    \toprule
    Model & Model Size & Link \\
    \midrule
    \midrule
     LLAMA \cite{touvron2023llama} & 7B & \url{https://huggingface.co/decapoda-research/llama-7b-hf} \\
     & 13B & \url{https://huggingface.co/decapoda-research/llama-13b-hf} \\
     \midrule
     LLAMA2  \cite{touvron2023llama_2}    & 7B  & \url{https://huggingface.co/meta-llama/Llama-2-7b-hf} \\
      & 13B & \url{https://huggingface.co/meta-llama/Llama-2-13b-hf} \\
     \midrule
     Stanford Alpaca \cite{alpaca} & 7B & \url{https://huggingface.co/allenai/open-instruct-stanford-alpaca-7b} \\
     \midrule
     Alpaca-LoRA \cite{alpaca-lora} & 7B &  \url{https://huggingface.co/tloen/alpaca-lora-7b}\\
     \midrule
     Vicuna(FastChat) \cite{vicuna} & 7B & \url{https://huggingface.co/lmsys/vicuna-7b-v1.3} \\
     & 13B & \url{https://huggingface.co/lmsys/vicuna-13b-v1.3}\\
     \midrule
     StableLM-Tuned-Alpha \cite{stablelm} & 7B & \url{https://huggingface.co/CarperAI/stable-vicuna-13b-delta}\\
     \midrule
     Databricks Dolly-v2 \cite{dolly} & 7B & \url{https://huggingface.co/databricks/dolly-v2-7b} \\
     & 12B & \url{https://huggingface.co/databricks/dolly-v2-12b}\\
     \midrule
     ChatGLM \cite{du2022glm, zeng2023glm-130b} & 6B & \url{https://huggingface.co/THUDM/chatglm-6b}\\
     \midrule
     MOSS \cite{moss} & 16B & \url{https://huggingface.co/fnlp/moss-moon-003-sft}\\
     \midrule
     Open-Assistant \cite{openassistant} & 7B &\url{https://huggingface.co/OpenAssistant/stablelm-7b-sft-v7-epoch-3}\\
     \midrule
     Openchatkit \cite{openchatkit} & 7B & \url{https://huggingface.co/togethercomputer/Pythia-Chat-Base-7B}\\
     \midrule
     BELLE \cite{belle} & 7B & \url{https://huggingface.co/BelleGroup/BELLE-7B-2M}\\
     \midrule
     MPT \cite{mpt7b} & 7B & \url{https://huggingface.co/mosaicml/mpt-7b-instruct}\\
     \midrule
     PandaLM \cite{PandaLM} & 7B & \url{https://huggingface.co/WeOpenML/PandaLM-7B-v1}\\
     \midrule
     RWKV(Pile) \cite{peng2023rwkv} & 7B & \url{https://huggingface.co/sgugger/rwkv-7b-pile}\\
     \midrule
     h2oGPT \cite{h2ogpt} & 6.9B & \url{https://huggingface.co/h2oai/h2ogpt-oig-oasst1-512-6_9b}\\
     & 12B & \url{https://huggingface.co/h2oai/h2ogpt-oasst1-512-12b}\\
     \midrule
     RedPajama(Base) \cite{RedPajama-INCITE} & 7B & \url{https://huggingface.co/togethercomputer/RedPajama-INCITE-7B-Base}\\
     RedPajama(Instruct) \cite{RedPajama-INCITE} & 7B & \url{https://huggingface.co/togethercomputer/RedPajama-INCITE-7B-Instruct}\\
     \midrule
     Galactica \cite{DBLP:journals/corr/abs-2211-09085} & 6.7B & \url{https://huggingface.co/facebook/galactica-6.7b}\\
     \bottomrule
    \end{tabular}
\end{table*}

\begin{table*}[!ht]
    \centering
    \caption{Overview of our evaluated multimodal models.}
    \label{tab2:eval_models_MM}
    \begin{tabular}{p{0.45\columnwidth}cp{1.1\columnwidth}}
    \toprule
    Model & Model Size & Link\\
    \midrule
    \midrule
    MiniGPT4 \cite{zhu2023minigpt} & 7B & \multirow{2}{*}{\url{https://github.com/Vision-CAIR/MiniGPT-4}}\\
    & 13B & \\
    \midrule
    LLaVA \cite{liu2023visual} & 7B & \url{https://huggingface.co/liuhaotian/LLaVA-7b-delta-v0}\\
    & 13B & \url{https://huggingface.co/liuhaotian/LLaVA-7b-delta-v0}\\
    \midrule
    MultiModal-GPT~\cite{gong2023multimodal} & 7B & \url{https://github.com/open-mmlab/Multimodal-GPT}\\
    \midrule
    VPGTrans~\cite{zhang2023transfer}   &7B & \url{https://github.com/VPGTrans/VPGTrans}\\
    \midrule
    Otter~\cite{li2023otter}            &7B& \url{https://huggingface.co/luodian/OTTER-Image-MPT7B}\\
    \midrule
    mPLUG-Owl~\cite{ye2023mplug}        &7B& \url{https://huggingface.co/MAGAer13/mplug-owl-llama-7b}\\
    
    \bottomrule
    
    \end{tabular}

\end{table*}

\subsection{Evaluation Prompts}

\begin{figure*}
    \centering
\includegraphics[width=\linewidth]{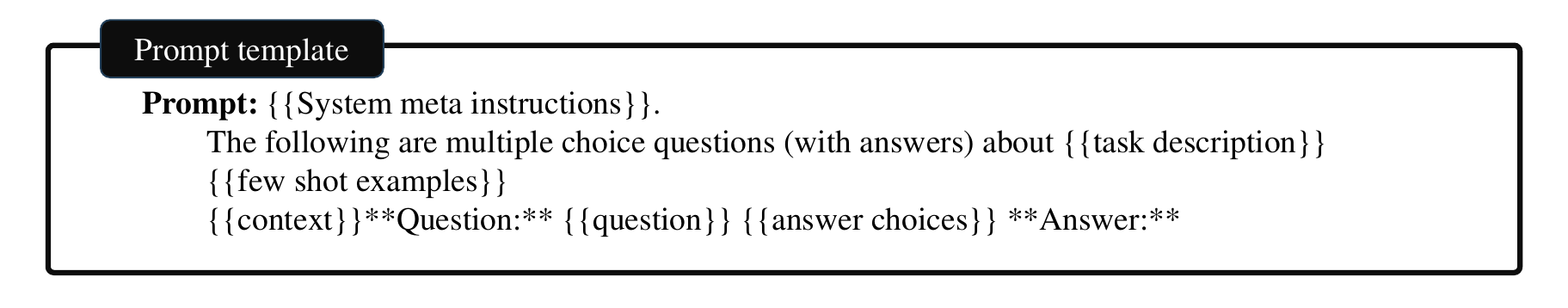}
    \caption{Prompt template used in our model evalution, following \cite{DBLP:journals/corr/abs-2303-13375}. You should replace content enclosed within double braces ${{}}$ with question-specific values.}
    \label{fig2:prompt_template}
\end{figure*}

To establish a baseline model's performance and enable fair comparisons, we adopt a similar prompting structure as utilized in the work by \cite{DBLP:journals/corr/abs-2303-13375} through our whole evaluation and study. The prompt template we use is illustrated in Fig. \ref{fig2:prompt_template}. Moreover, in Fig. \ref{fig2:prompt_template}, the placeholders $\{{\{System~meta~ instructions}\}\}$ will be substituted with corresponding text from Table \ref{tab2:sys_meta_inst} for models that have officially published system meta instructions. For models without such instructions, we use a null character string to replace the placeholders. Moreover, following \cite{DBLP:journals/corr/abs-2303-13375}, we leverage the \textit{logit\_bias} to encourage the model to produce exclusively valid responses.

\begin{table*}
\centering
\caption{System meta instructions for each model (if any).}
\label{tab2:sys_meta_inst}
\resizebox{!}{!}{
\begin{tabular}{lp{1.2\columnwidth}}
\toprule
Model & System meta instruction \\
\midrule
ChatGLM\cite{du2022glm, zeng2023glm-130b}    &    "Below is an instruction that describes a task. Write a response that appropriately completes the request.$\backslash$n \#\#\# Human: " + query + $\backslash$n + "\#\#\# Assistant:"          \\
\midrule
Alpaca \cite{alpaca, alpaca-lora}      &  "Below is an instruction that describes a task. Write a response that appropriately completes the request.$\backslash$n$\backslash$n \#\#\# Instruction:$\backslash$n" + query + "$\backslash$n$\backslash$n" + "\#\#\# Response:$\backslash$n" \\
\midrule
Vicuna \cite{vicuna}      &     "A chat between a curious user and an artificial intelligence assistant. The assistant gives helpful, detailed, and polite answers to the user's questions. " + "Question: " + query + " " + "ASSISTANT:"                                  \\
\midrule
StableLM-Tuned-Alpha \cite{stablelm}      &    """<|SYSTEM|>\# StableLM Tuned (Alpha version) \par
- StableLM is a helpful and harmless open-source AI language model developed by StabilityAI.\par
- StableLM is excited to be able to help the user but will refuse to do anything that could be considered harmful to the user.\par
- StableLM is more than just an information source, StableLM is also able to write poetry, short stories, and make jokes.\par
- StableLM will refuse to participate in anything that could harm a human.\par
""" + "<|USER|>" + query + "<|ASSISTANT|>"                    \\
\midrule
Databricks Dolly-v2 \cite{dolly} & "Below is an instruction that describes a task. Write a response that appropriately completes the request.$\backslash$n$\backslash$n\#\#\# Instruction:$\backslash$n" + query + "$\backslash$n$\backslash$n" + "\#\#\# Response:$\backslash$n" \\
\midrule
MOSS \cite{moss} & "You are an AI assistant whose name is MOSS.$\backslash$n- MOSS is a conversational language model that is developed by Fudan University. It is designed to be helpful, honest, and harmless.$\backslash$n + "<|Human|>: " + query + "<eoh>$\backslash$n<|MOSS|>:" \\
\midrule
Open-Assistant \cite{openassistant} & "Below is an instruction that describes a task. Write a response that appropriately completes the request.$\backslash$n$\backslash$n |prompter|> + query + <|endoftext|><|assistant|>"  \\
\midrule
Openchatkit \cite{openchatkit} & "Below is an instruction that describes a task. Write a response that appropriately completes the request.$\backslash$n$\backslash$n <human>:  + query + $\backslash$n<bot>:"\\
\midrule
BELLE-7B \cite{belle} & "Below is an instruction that describes a task. Write a response that appropriately completes the request.$\backslash$n$\backslash$n Human:  + query + $\backslash$n$\backslash$nBelle: " \\
\midrule
MPT \cite{mpt7b} & "Below is an instruction that describes a task. Write a response that appropriately completes the request.$\backslash$n$\backslash$n \#\#\# Instruction: + query + $\backslash$n$\backslash$n\#\#\# Response:" \\
\bottomrule
\end{tabular}
}
\end{table*}

\section{Pre-Training and Fine-tuning Data }

Pre-training and fine-tuning are essential steps in the development of GPT and other language models. In this section, we explore the key aspects of different data utilized during these stages, examining data quality, diversity, quantity, and sources for pre-training. Additionally, we investigate the various types of fine-tuning data, such as the instructions and alignment data used to tailor GPT models to specific tasks and domains.

\subsection{Pre-training Data}
Pre-training involves training the GPT model on a large corpus of unlabeled text to learn general language representations. The data used during pre-training is important since it will significantly impact the model's performance and adaptability.

\subsubsection{Data Source}
When talking about data sources, we refer to the public and private data used in the pre-training stage. 

\paratitle{Public Data}. Public and open-sourced datasets play a vital role in pre-training GPT models, offering a diverse and vast collection of text data from various domains. Leveraging publicly available corpora, such as Wikipedia \cite{Wikipedia}, Common Crawl \cite{commoncrawl}, BooksCorpus \cite{zhu2015aligning} and the Pile \cite{gao2020pile} (more data can be found in Table~\ref{tab2:stat_models}), provides access to a rich source of language data that enables models to learn general language representations. However, using public data also comes with limitations: the quality of data can be inconsistent, and noise or inaccuracies may be present \cite{brown2020language}. Pre-training on noisy data can impact model performance, leading to undesirable biases or irrelevant associations in the learned representations \cite{brown2020language}.

\paratitle{Private Data}. The use of proprietary or custom data collections for pre-training GPT models offers unique advantages and challenges. By leveraging private datasets specific to a particular organization or domain, models can be fine-tuned to cater precisely to the needs and requirements of the intended applications. Proprietary data may include sensitive or proprietary information, making it unsuitable for public release but valuable for enhancing model performance within the organization. Several platforms like Lamini \cite{lamini} provide opportunities for users to train custom models on private data. In another view, first pre-training on public data and then continue pre-training on the private data is also a common choose. 

\subsubsection{Data Quantity and Quality}
Pre-training typically requires a vast amount of data to achieve high performance, especially when the LLM parameter scale increases dramatically \cite{touvron2023llama, brown2020language, DBLP:journals/corr/abs-2203-15556}. Larger datasets enable models to learn more robust and nuanced language representations. Along the history of the large language model development, we can see that the data scale increased significantly, same as the model parameters, for example, from billion tokens to tens of trillion tokens. However, it is crucial to emphasize that solely high-quality data are essential for performance improvement. Merely augmenting the dataset with low-quality data can result in a decline in performance \cite{DBLP:journals/corr/abs-2205-10487}. Therefore, recently, there is another trend that obtaining high quality data instead of tremendous scale of low quality data for pre-training. 

Since the quality of pre-training data significantly impacts the effectiveness of GPT models and ensures the models capture accurate and reliable information during pre-training, we need to ensure data excellence by involving carefully curating and filtering the text corpus to remove noise, errors, and irrelevant content, ultimately leading to more robust and reliable GPT models \cite{DBLP:conf/icml/DuHDTLXKZYFZFBZ22, DBLP:journals/corr/abs-2305-13169}. Next, we will introduce some data preprocessing strategies for data quality improvement in several representative LLMs, including classifier-based filtering, rule-based filtering and utilizing textbook quality data.

\paratitle{Classifier-based Filtering}. In this approach, a selection classifier is trained using high-quality texts as examples. This classifier is then used to analyze the remaining data and distinguish between high-quality and low-quality samples. The low-quality data is subsequently eliminated from the dataset, ensuring that only the most reliable and valuable information is retained for further pre-training.

\paraitem{GPT-3} \cite{brown2020language} devised an automated filtering method to effectively eliminate low-quality documents from the Common Crawl dataset. It used a classifier trained on high-quality data (WebText, Wikipedia, and web books corpus) to prioritize documents with higher scores, resulting in improved data quality for generative text samples. Moreover, GPT-3 adopted fuzzy deduplication of documents within each dataset and the removal of WebText from Common Crawl, further improving the data quality.

\paraitem{PaLM} \cite{chowdhery2022palm}, following \textbf{GLaM} \cite{DBLP:conf/icml/DuHDTLXKZYFZFBZ22}, developed a text quality classifier to distinguish between high-quality content (curated text from Wikipedia, books, and selected websites) and other webpages. Using a feature hash-based linear classifier, they both estimated the content quality of webpages and applied a Pareto distribution to sample webpages based on their scores, including some lower-quality webpages to avoid biases in the classifier.

\paratitle{Rule-based Filtering}. In this method, heuristic-based approaches are utilized to identify and eliminate low-quality texts based on a set of well-designed rules. These rules are carefully crafted to detect specific patterns or characteristics that are indicative of low-quality content. By applying these rules to the dataset, low-quality texts can be effectively filtered out, ensuring that only high-quality and reliable data remains for further processing.

\paraitem{Gopher} \cite{DBLP:journals/corr/abs-2112-11446} uses simple heuristic filters to remove low-quality text based on word count, word lengths, symbol-to-word ratios, bullet points, and ellipses. They ensure data quality by filtering out documents lacking essential English words and removing exact and near-duplicate documents.

\paraitem{BLOOM} \cite{scao2022bloom} aimed to identify high-quality text written by humans for humans, excluding non-natural language content like preprocessing errors, SEO pages, or spam. It defined a set of quality indicators individually tailored for each language, with parameters and thresholds chosen by fluent speakers. Additionally, the authors manually reviewed each source to identify the indicators best suited to detect non-natural language, and visualization tools supported these processes.

\paraitem{Falcon LLM} \cite{falcon, DBLP:journals/corr/abs-2306-01116} is a language model pretrained on the RefinedWeb Dataset \cite{DBLP:journals/corr/abs-2306-01116}, an English web-only pretraining dataset containing five trillion tokens. The authors claimed that by using properly filtered and deduplicated web data, Falcon LLM achieves impressive performance, even surpassing state-of-the-art models trained on The Pile. The filtering process involves both document-wise and line-wise procedures, using heuristics to detect and eliminate excessive repetitions and non-natural language content.

\paratitle{Textbook-Quality Data Curation}. This method focuses on curating a dataset comprising high-quality texts akin to comprehensive textbooks. The aim is to ensure the training data comprises clear, well-structured, and informative examples, which ultimately enhances language models' proficiency. By using such textbook-quality data, the resulting dataset significantly improves the models' learning efficiency and performance across various tasks.

\paraitem{phi-1-base} \cite{DBLP:journals/corr/abs-2306-11644} is a large language model for code, pre-trained on a curated collection of "textbook quality" data extracted from the web (6B tokens) and supplemented with synthetically generated textbooks. Leveraging such "textbook quality" data, phi-1-base achieves remarkable performance on coding benchmarks, surpassing many open-source models, despite its smaller model and dataset sizes. The authors propose that the inclusion of such high-quality data enhances the learning efficiency of language models for code, as it provides clear, self-contained, instructive, and well-balanced examples of coding concepts and skills.

\subsubsection{Data Diversity}
The diversity of data used during pre-training is essential to enhance the model's ability to generalize across various tasks and domains. Incorporating a diverse range of text sources, such as news articles, books, social media data, code, and advanced science
and math journals has been shown to enrich the model's understanding of language \cite{DBLP:journals/corr/abs-2305-13169, DBLP:journals/corr/abs-2112-11446}. By training on diverse dataset, the model is able to gain knowledge from different aspects, and ease the downstream finetuning. Most of the LLMs are trained with a mixture of data from diverse sources \cite{DBLP:journals/corr/abs-2112-11446, black2022gpt, chowdhery2022palm}. Table \ref{tab2:stat_models} presents the training data source for some representative LLMs.


\subsection{Fine-tuning / Instruction Tuning Data}
Fine-tuning, and instruction tuning, is the process of adapting the pre-trained GPT models to perform specific tasks. During this stage, GPT models are further trained on task-specific datasets, and various fine-tuning instructions are employed to guide the model's behavior.

\subsubsection{Human Instruction Data}
Human instruction involves providing explicit task-specific instructions designed by human to guide the model's responses. This approach is common in applications where clear task guidelines are available. 

\paratitle{Human-written Data}. Human-written data consists of carefully crafted instructions and prompts designed by human annotators. This data is manually curated to ensure high quality and relevance to the target tasks.

\paraitem{InstructGPT} \cite{ouyang2022training}. In InstructGPT, human-written data is used to train the initial models. Labelers were asked to create three types of prompts: plain prompts with arbitrary tasks to ensure diversity, few-shot prompts with instructions and multiple query/response pairs, and user-based prompts corresponding to specific use cases. These prompts were then used to create three datasets: SFT dataset for training SFT models, RM dataset for training reward models with labeler rankings, and PPO dataset for RLHF fine-tuning without human labels. The SFT dataset contains about 13k training prompts, the RM dataset has 33k training prompts, and the PPO dataset has 31k training prompts from the API.

\paratitle{ShareGPT Data}. ShareGPT is a user-friendly Chrome Extension that simplifies sharing ChatGPT conversations effortlessly \footnote{\url{https://sharegpt.com}.}. Some datasets can also be obtained from ShareGPT, offering a valuable resource for researchers and enthusiasts interested in utilizing ChatGPT's conversational data. 

\paraitem{ShareGPT-90K (formerly 52k)} \cite{sharegpt52K}. This dataset comprises around 90k (52k in the old version) conversations obtained via the ShareGPT API. These conversations encompass user prompts and responses from OpenAI's ChatGPT.

\paraitem{ShareGPT-Vicuna-70k} \cite{vicuna}. This dataset comprises approximately 70k user-shared conversations obtained from the ShareGPT API. To ensure high data quality, the dataset undergoes a process of converting HTML to markdown and filtering out inappropriate or low-quality samples. Additionally, to accommodate the model's context length, lengthy conversations are divided into smaller segments. However, due to various concerns, the authors have not released the dataset \footnote{\url{https://github.com/lm-sys/FastChat/issues/181}.}.

\subsubsection{Synthesis Instruction Data} 
Synthesis instruction data serves as a valuable resource for fine-tuning GPT models when explicit task-specific instructions are scarce or unavailable. In scenarios where human-crafted instructions are challenging and costly to obtain, synthesis instruction provides a creative solution to generate instruction data. By using GPT models themselves, or other language models, to synthesize inputs and corresponding outputs (instruction pair), this technique creates a synthetic dataset for fine-tuning. Next, we will present two types of synthesis instruction datasets.

\paratitle{Self-Instructed Data}. The Self-Instruct framework \cite{selfinstruct} is a groundbreaking approach that enables language models to enhance their understanding and adherence to natural language instructions. Utilizing the model's generated responses, Self-Instruct creates a substantial collection of instructional data, leading to significant improvements in the model's ability to follow instructions without the need for labor-intensive manual annotation.

\paraitem{Self-Instruct-52k} \cite{selfinstruct} is a synthesis instruction dataset generated by the Self-Instruct framework \cite{selfinstruct}, containing 52k instructions and over 82k instances associated with these instructions. In \cite{selfinstruct}, Self-Instruct-52k is generated by GPT-3, i.e., “\textit{davinci}” engine of OpenAI API.

\paraitem{Stanford Alpaca-52k} \cite{alpaca} followed the data synthesis pipeline from \cite{selfinstruct} and made the main modifications as the following: (1) replaced “\textit{davinci}” engine with "\textit{text-davinci-003}" for instruction data generation; (2) made batch decoding more aggressive, generating 20 instructions simultaneously, significantly reducing data generation costs; (3) simplified the pipeline by disregarding the classification/non-classification instruction difference. Stanford Alpaca-52k also consists of 52k generated data but is much more diverse than Self-Instruct-52k.

\paraitem{GPT-4 English/Chinese Instruction-Following Data} \cite{DBLP:journals/corr/abs-2304-03277} is an extension of the Stanford Alpaca-52k dataset \cite{alpaca}, comprising 52k English instruction-following samples and 52k Chinese instruction-following samples. In contrast to the "\textit{text-davinci-003}" engine used in the original dataset, the authors utilized the "\textit{gpt-4}" engine for data generation. For the English instruction data, GPT-4 generates corresponding English responses. Meanwhile, for the Chinese instruction data, ChatGPT assists in translating the 52k instructions into Chinese, followed by GPT-4 generating the corresponding answers in Chinese.

\paratitle{Converted Datasets into Instructional Format}. Since creating an instruction-tuning dataset with many tasks from scratch would be resource-intensive, researchers start exploring the possibility of acquiring data from existing datasets. That is, existing datasets from the research community are adapted and reformatted into an instructional format. This process converts the original data into a structure that provides clear instructions for each task, making it suitable for instruction-based fine-tuning of language models.

\paraitem{Flan 2021 Dataset} \cite{DBLP:conf/iclr/WeiBZGYLDDL22}. This dataset is created by transforming existing publicly available text datasets into an instructional format for instruction tuning\footnote{\url{https://huggingface.co/datasets/conceptofmind/flan2021_submix_original}.}. It consists of 62 datasets, categorized into twelve task clusters. For each dataset, ten unique templates with natural language instructions are manually composed. The pre-trained language model is instruction-tuned on this mixture of datasets using randomly selected instruction templates. The goal is to improve the model's ability to follow specific guidelines and perform task-oriented behaviors effectively.

\paraitem{Flan Collection} \cite{DBLP:journals/corr/abs-2301-13688}. The Flan Collection compiles various datasets and data augmentation methods for instruction tuning. It includes datasets from \textbf{Flan 2021} \cite{DBLP:conf/iclr/WeiBZGYLDDL22}, \textbf{P3} \cite{sanh2021multitask}, \textbf{Super-Natural Instructions} \cite{DBLP:conf/emnlp/WangMAKMNADASPK22}, and others, formatted into zero-shot, few-shot, and chain-of-thought templates. The dataset is organized into sub-mixtures, each with different variations of prompts, including answer options or not. It contains 1,836 finetuning tasks by combining the mixtures from prior work. Flan Collection serves as a valuable resource for instruction-based fine-tuning and achieves strong performance on evaluation benchmarks with Flan-T5 \cite{chung2022scaling} and Flan-PaLM \cite{chung2022scaling} models.

\subsubsection{Alignment Data}

Alignment data involves data with inputs and corresponding outputs, providing direct supervision for fine-tuning. While the model's foundational knowledge and capabilities stem from pretraining, its true efficacy and user interactions are refined through alignment data, instructing it in the most effective subdistribution of formats for engaging with users \cite{DBLP:journals/corr/abs-2306-11644}. Two primary approaches to leverage alignment data are reinforcement learning from human feedback (RLHF) \cite{ouyang2022training, openai2023gpt} and supervised fine-tuning \cite{DBLP:journals/corr/abs-2306-11644}.

\paratitle{Reinforcement Learning from Human Feedback (RLHF)}. RLHF is a machine learning technique that utilizes human-provided feedback to train models through reinforcement learning. The process involves two key steps: (1) Collecting manually ranked comparison response pairs to build a reward model for evaluating the quality of generated responses. (2) Optimizing the model (policy) using the reinforcement learning framework with rewards obtained from the trained reward model. RLHF enables models to improve their performance based on human feedback, making it a valuable approach for enhancing language generation tasks.

\paraitem{InstructGPT} \cite{ouyang2022training}. InstructGPT utilizes reinforcement learning from human feedback (RLHF) to fine-tune GPT-3 based on human preferences. The data is labeled by a team of 40 contractors, collecting demonstrations of desired output behavior on prompts from the OpenAI API, generating approximately 33k samples for RLHF. A reward model (RM) is trained on human-labeled comparisons between model outputs, and the PPO algorithm is employed for fine-tuning, resulting in the aligned InstructGPT.

\paraitem{GPT-4 Comparison Data} \cite{DBLP:journals/corr/abs-2304-03277}. The GPT-4 Comparison data consists of ratings provided by GPT-4 for its own responses on a scale from 1 to 10. Additionally, GPT-4 is tasked with comparing and rating responses from three models: GPT-4, GPT-3.5, and OPT-IML \cite{DBLP:journals/corr/abs-2212-12017}. These ratings serve as training data for reward models in the RLHF process.

\paratitle{Supervised Fine-tuning}. Supervised fine-tuning for LLM alignment is also a powerful technique that enables the alignment of language models with specific instructional requirements through targeted data augmentation. By leveraging human-provided feedback and carefully curated datasets, this approach fine-tunes LLMs to better follow instructions and produce contextually relevant responses.

\paraitem{LIMA} \cite{DBLP:journals/corr/abs-2306-11644}. LIMA is trained on a dataset of 1,000 prompts and responses, stylistically aligned in the manner of a helpful AI assistant. The data is curated from multiple sources, including community Q\&A forums like Stack Exchange, wikiHow, and Pushshift Reddit Dataset, as well as manually authored examples. In comparison to DaVinci003 and Alpaca, LIMA demonstrates superior performance in a human preference study, with only a small gap compared to GPT-4 and Claude. These results reinforce the notion that large language models predominantly acquire their knowledge during pretraining, while limited instruction tuning data suffices to achieve high-quality output. Moreover, the authors examined data quality, quantity, and diversity. Expanding input prompt diversity and enhancing output quality have positive impacts, while increasing data quantity may not yield the same benefits. Comparing various datasets, more diverse prompts led to significantly higher performance. Filtering data for quality improvement also showed positive results. Doubling the training set did not improve response quality.


\subsubsection{Domain-specific Data}
In domain-specific applications, fine-tuning with domain-specific data is crucial for achieving optimal performance. Researchers have shown the importance of fine-tuning GPT models with domain-specific datasets to tailor the models to specific industries or tasks. 

\paratitle{Math Domain}. Fine-tuning GPT models with math-related datasets is vital to enhance their capabilities in solving mathematical problems and providing accurate mathematical expressions. Math data aids in developing language models that can effectively comprehend and generate mathematical content, enabling applications in educational settings, scientific research, and various technical fields.

\paraitem{Goat-Data}~\cite{liu2023goat}. Goat is a model finetuned for arithmetic tasks on LLaMA. To improve the mathematical capability of language models, such as solving challenging tasks, large-number multiplication and division, Goat splits the tasks into learnable simple tasks, and performs a chain-of-thought learning. Therefore, they create a dataset that contains instruction data with arithmetic expressions in random templates. The dataset is released in Github\footnote{\url{https://github.com/liutiedong/goat.}}.

\paraitem{PRM800K} \cite{DBLP:journals/corr/abs-2305-20050}. The PRM800K Dataset comprises 800K step-level labels, obtained exclusively from a large-scale generator. It includes solutions to 12K problems, totaling 75K solutions. Trained with RLHF on PRM800K, the model in \cite{DBLP:journals/corr/abs-2305-20050} solve 78.2\% of problems from a representative subset of the MATH test set \cite{DBLP:conf/nips/HendrycksBKABTS21}.

\paratitle{Scientific Domain}. For scientific domains such as biology, chemistry, or physics, fine-tuning GPT models with scientific datasets is crucial. This process empowers the models to grasp scientific jargon, comprehend complex concepts, and generate contextually relevant scientific content. Fine-tuned GPT models can assist researchers in automating literature review tasks, suggesting hypotheses, and generating scientific reports with accuracy and domain-specific context. 

\paraitem{Filtered S2ORC} \cite{DBLP:journals/corr/abs-2304-14454}. The dataset is used for finetuning PMC-LLaMA \cite{DBLP:journals/corr/abs-2304-14454}. The dataset starts with the S2ORC Datasets, consisting of 81.1M English-language academic papers. After filtering them using PubMed Central (PMC)-id, approximately 4.9M papers remain, focusing on medical knowledge and containing over 75B tokens.

\paratitle{Code Domain}. In the field of software development, fine-tuning GPT models with code-related datasets holds immense value. By leveraging code-specific data, language models can better understand programming languages, code syntax, and logic, enabling them to assist developers in code completion, bug detection, and code summarization tasks. Fine-tuned GPT models in the code domain contribute to increased developer productivity and improved code quality.

\paraitem{CodeExercises} \cite{DBLP:journals/corr/abs-2306-11644}. The CodeExercises dataset, extensively employed in the development of the powerful phi-1 model \cite{DBLP:journals/corr/abs-2306-11644}, constitutes a relatively compact yet valuable collection of Python exercises and solutions, comprising less than 180 million tokens. Each exercise represents a function that necessitates completion, presented in the form of a docstring. The primary focus of this dataset lies in aligning the phi-1 model's capabilities to excel at function completion tasks based on natural language instructions.

\section{Deployment and Finetuning Techniques}
\subsection{Efficient Deployment} 
Most foundation models are typically trained in the FP16/BF16 format, which offers nearly twice the efficiency of FP32 training. However, they still demand a significant amount of GPU memory during deployment, making them unsuitable for certain low-resource scenarios.

Quantization refers to the process of minimizing the number of bits used to represent numerical values. This technique brings several advantages, including reduced model size, lower memory requirements, and diminished computational demands. Over the years, various strategies for quantizing large models have gained considerable traction. Here, we briefly introduce some of these techniques.

\textbf{ZeroQuant}~\cite{yao2022zeroquant} is designed for zero-cost compression to INT8 precision of weights and activations. It utilizes a hardware-friendly quantization scheme, an interlayer knowledge distillation algorithm, and a highly optimized quantization system backend.

\textbf{LUT-GEMM}~\cite{park2022nuqmm} focuses on model size reduction by quantizing weights using a non-uniform quantization method. It also accelerates quantized matrix multiplications using a novel kernel. The approach allows for a wide trade-off between compression ratio and accuracy, resulting in a significant acceleration of inference speed and reduced energy consumption.

\textbf{LLM.int8$()$}~\cite{dettmers2022llm} enables 8-bit matrix multiplication in Transformer models, halving the memory required for inference without sacrificing performance. The authors demonstrate that their technique permits the use of large language models with up to 175B parameters without any performance degradation.

Compared to these approaches, \textbf{GPTQ}~\cite{frantar2022gptq} adopts a layer-wise quantization strategy, solving a corresponding reconstruction problem for each layer. It builds upon the Optimal Brain Quantization (\textbf{OBQ}) method, with significant modifications to make it scalable for large language models. The method applies quantization arbitrarily, updates weights lazily in batches, and uses a Cholesky reformulation to address numerical inaccuracies.

Another technique, \textbf{SmoothQuant}~\cite{xiao2023smoothquant}, proposes an accurate and efficient post-training quantization method for large language models. By smoothing the outliers of activation values, it shifts the quantization difficulty from activations to weights, enabling 8-bit quantization of both weights and activations. As a result, it achieves high speed and reduced memory usage with almost no loss in accuracy.

In addition to post-training quantization methods, \textbf{LLM-QAT}~\cite{liu2023llmqat} investigates quantization-aware training. It utilizes the pretrained full-precision model as a teacher model to generate training data for the quantized student model. The predictions of the pretrained full-precision model are utilized to distill knowledge into the quantized student model.

\subsection{Efficient Finetuning}
The most common and straightforward way to adapt foundation models to downstream tasks is by finetuning downstream task data. However, finetuning the whole model parameters is still energy-consuming and requires a large GPU memory. Parameter-efficient finetuning aims to only finetune a small amount of the parameters while maintaining comparable performance to full parameter fine-tuning.

\textbf{Adapter Tuning.} Adapter Tuning~\cite{houlsby2019parameter} is a technique in deep learning that allows for quicker and more efficient adaptation of pre-trained models to new tasks. The technique involves adding small, task-specific "adapter" modules (e.g. feedforward layers with skip-connections), which are lightweight neural networks that can be plugged into pre-trained models to fine-tune them for specific tasks. The weights of the adapters are then trained on the new task, while the weights of the pre-trained model are frozen. This allows for efficient transfer learning, as only a small number of parameters need to be updated.

\textbf{Low-Rank Adaption (LoRA).} LoRA freezes the pre-trained model weights and injects trainable rank decomposition matrices into each layer of the transformer architecture~\cite{hu2021lora}. More specifically, a pre-trained weight matrix $W_0 \in \mathbb{R}^{d \times k}$ is updated by a low-rank decomposition $W_0 + \Delta W = W_0 + BA$, where $B \in \mathbb{R}^{d\times m}, A \in \mathbb{R}^{m\times k}$, and $m\ll min(d, k)$. Only $A$ and $B$ are trainable parameters during finetuning. Recently, QLoRA~\cite{dettmers2023qlora} proposes to quantize
a pre-trained model to 4-bit and incorporates a limited number of learnable Low-rank Adapter weights. It significantly decreases the average memory needs for finetuning a 65-billion-parameter model from over 780GB of GPU memory to less than 48GB, while maintaining the runtime and predictive accuracy comparable to a fully finetuned 16-bit baseline. (Note that some recent works also explore low-rank updates during pretraining, such as ReLoRA~\cite{lialin2023stack}.)

\textbf{Continuous Prompt Tuning and Prefix Tuning.}
Continuous prompt tuning prepends or inserts learnable prompts to input sequence and freezes the pre-trained model weights~\cite{lester2021power, liu2021gpt}. It is shown that continuous prompt tuning is comparable to finetuning on simple classification tasks with 10-billion-parameter models. Prefix tuning\cite{li2021prefix} prepends prefixes to Transformer (more specifically, every Transformer layer has trainable continuous prompts rather than merely the input layer) and achieves comparable performance in table-to-text generation tasks compared with full parameter fine-tuning. Further empirical evidence from Ptuning-v2~\cite{liu2021p} demonstrates that prefix tuning achieves comparable performance to finetuning across different scales and tasks.

\section{Open-source Tools}

In this section, we introduce some popular open-sourced projects developed for better large language model ecology, e.g., model training and deployment, efficient finetuning and open leaderboard. An overview can be seen in Table \ref{tab:4_eff_tools}.

\begin{table*}[!ht]
\centering
\caption{Overview of open-source efforts and tools development for large language models}
\label{tab:4_eff_tools}
\scalebox{0.9}{
\begin{tabular}{ccp{0.5 \columnwidth}cp{0.65 \columnwidth}}
\toprule
Tool & Category & Application & Released by & Link \\
\midrule
 Transformers \cite{wolf2020transformers} & Deployment &    LLM training and deployment     &       Huggingface      &   \url{https://huggingface.co/transformers}   \\
 Colossal-AI \cite{li2021colossal} & 
 Deployment &Unified system to train and deploy large-scale models & HPC-AI Tech & \url{https://colossalai.org/} \\
 GPT4all \cite{gpt4all} & Deployment & Large and personalized language models training and deployment on common hardware & Nomic AI & \url{https://gpt4all.io/} \\
 PandaLM \cite{PandaLM} & Deployment & System providing automated and reproducible comparisons among various LLMs & Westlake University & \url{https://github.com/WeOpenML/PandaLM} \\
 MLC LLM \cite{mlcllm} & Deployment & Solution allowing LLMs to be deployed natively & MLC AI & \url{https://mlc.ai/mlc-llm/}\\
 \midrule
 Deepspeed \cite{rasley2020deepspeed} & Accelerating    &    Accelerating training and inference of large-scale models    &    Microsoft         &   \url{https://github.com/microsoft/DeepSpeed}   \\
 Megatron-LM  \cite{shoeybi2019megatron,narayanan2021efficient,korthikanti2023reducing}  & Accelerating &     Accelerating training and inference of large-scale models    &       Nvidia      &  \url{https://github.com/NVIDIA/Megatron-LM}   \\
 \midrule
 MinGPT \cite{mingpt} & Reproduction & Re-implementation of GPT which is clean, interpretable and educational & Stanford University & \url{https://github.com/karpathy/minGPT}\\
 RedPajama \cite{RedPajama} & Reproduction & An effort to produce reproducible and fully-open language models & ETH Zurich & \url{https://together.xyz/blog/redpajama}\\
 \midrule
 LangChain \cite{langchain} & Framework & Framework for integration of LLMs with other computational sources and knowledge & LangChain & \url{https://python.langchain.com/}\\
 xTuning \cite{xturing} & Framework & Framework providing fast, efficient and simple fine-tuning of LLMs & Stochastic  & \url{https://github.com/stochasticai/xturing}\\
 Open LLM Leaderboard \cite{open-llm-leaderboard} & Evaluation & LM evaluation leaderboard & Huggingface & \url{https://huggingface.co/spaces/HuggingFaceH4/open_llm_leaderboard}\\
 Scikit-LLM \cite{scikit-llm} & Framework & Framework integrating LLMs into scikit-learn for enhanced text analysis tasks & Tractive & \url{https://github.com/iryna-kondr/scikit-llm}\\
 AlpacaFarm \cite{dubois2023alpacafarm} &Framework &Simulation framework for methods that learn from human feedback  & Stanford & \url{https://github.com/tatsu-lab/alpaca_farm/}\\
 h2oGPT \cite{h2ogpt} & Framework &LLM finetuning framework and chatbot UI with document(s) question-answer capabilities & H2O.ai & \url{https://github.com/h2oai/h2ogpt}\\
 \midrule
 Open-Assistant \cite{openassistant} & Software & Customized and personalized chat-based assistant & 
 LAION AI & \url{https://github.com/LAION-AI/Open-Assistant} \\
MetaGPT \cite{metagpt} & Software & Multi-agent framework to tackle tasks with multiple agents & 
 Open-Source Community & \url{https://github.com/geekan/MetaGPT} \\
 \midrule
 {PEFT} \cite{peft} & Finetuning & Library for finetuning LLMs with only part of parameters & Huggingface & \url{https://huggingface.co/docs/peft}\\
\bottomrule
\end{tabular}
}
\end{table*}

\textbf{Transformers}~\cite{wolf2020transformers} is an open-sourced deep learning library developed by Huggingface that provides advanced natural language processing (NLP) capabilities. In particular, the library comprises multiple pre-trained models based on Transformer architectures. The library has gained popularity in the NLP community due to its simplicity, flexibility, and the pre-trained models' ability to perform well across different tasks.

\textbf{Deepspeed}~\cite{rasley2020deepspeed} is a deep learning optimization library developed by Microsoft that helps accelerate large-scale AI models. It is an open-sourced library designed to optimize large GPT-style language models with up to billions of parameters. DeepSpeed offers a series of optimization techniques for both model training and inference, such as model parallelism, mixed-precision training, and pipeline parallelism. It also provides a set of tools and configuration options to simplify the process of setting up and tuning large-scale models.

\textbf{Megatron-LM}~\cite{shoeybi2019megatron,narayanan2021efficient,korthikanti2023reducing} is an open-sourced library developed by Nvidia for training large transformer language models at scale. It contains the model architecture and the training pipeline needed to create state-of-the-art transformer-based language models, including GPT, BERT, and T5.

\textbf{Colossal-AI}~\cite{li2021colossal} is an open-sourced deep learning framework developed by  HPC-AI Tech that provides a unified system to train and deploy large-scale models. It is designed to be easy to use and efficient, and it supports a variety of parallelism methods. \textbf{ColossalChat} is a free and publicly accessible alternative to ChatGPT that incorporates a comprehensive RLHF pipeline based on Colossal-AI.

\textbf{GPT4all}~\cite{gpt4all} is an open-sourced software ecosystem that is available for public use and enables individuals to train and utilize large and personalized language models on common hardware. The GPT4All software is designed to efficiently perform inference tasks for large language models with 7-13 billion parameters on the CPUs of laptops, desktops, and servers.

\textbf{MinGPT}~\cite{mingpt} is a PyTorch re-implementation of GPT, both training and inference. minGPT tries to be small, clean, interpretable, and educational. \textbf{NanoGPT}~\cite{nanogpt} is a rewrite of minGPT that prioritizes teeth over education. 

\textbf{RedPajama}~\cite{RedPajama} is an effort to produce reproducible and fully open language models. The project has released \textbf{RedPajama-Data}~\cite{redpajama-data} and \textbf{RedPajama-INCITE}~\cite{RedPajama-INCITE}. RedPajama-Data is an open-sourced project to reproduce the LLaMA training dataset and RedPajama-INCITE is a series of base models and instruction-tuned models.

\textbf{PandaLM}~\cite{PandaLM} aims to offer automated and reproducible comparisons among various large language models (LLMs). By providing PandaLM with identical context, it can evaluate the reactions of diverse LLMs and provide an explanation for the outcome, along with a reference answer. In contrast to ChatGPT, PandaLM is accessible to the public, reproducible, and guarantees data security.

\textbf{Open-Assistant}~\cite{openassistant}  is a chat-based assistant that comprehends tasks, interacts with external systems, and dynamically retrieves the information to accomplish them. It can be conveniently customized and personalized and is created as free, open-source software.

\textbf{LangChain}~\cite{langchain} is to facilitate the integration of large language models with other computational sources and knowledge. LangChain is tailored to assist in six key areas, namely LLMs and Prompts, Chains, Data Augmented Generation, Agents, Memory, and Evaluation.


\textbf{MLC LLM}~\cite{mlcllm} is a universal solution that allows any language models to be deployed natively on a diverse set of hardware backends and native applications, plus a productive framework for everyone to further optimize model performance for their own use cases.

\textbf{xTuning}~\cite{xturing} is a tool that offers a swift, effective, and user-friendly way to fine-tune LLMs. With its intuitive interface, xTuring simplifies the process of constructing, customizing, and managing LLMs by allowing users to fine-tune LLMs to their own data and applications. The entire process can be carried out within the user's computer or private cloud, thus ensuring data privacy and security.

\textbf{Open LLM Leaderboard} \cite{open-llm-leaderboard} is a language model evaluation leaderboard maintained by Huggingface. It can automatically evaluate models as long as it's a Huggingface transformers model with weights on the Hub. The evaluation datasets include ARC~\cite{Clark2018ThinkYH}, HellaSwag~\cite{zellers2019hellaswag}, MMLU~\cite{hendrycks2020measuring} and TrurhfulQA~\cite{lin2021truthfulqa}.

\textbf{Scikit-LLM}~\cite{scikit-llm} effortlessly incorporate powerful language models, such as ChatGPT, into scikit-learn~\cite{pedregosa2011scikit} to upgrade text analysis procedures.

\textbf{AlpacaFarm}~\cite{dubois2023alpacafarm} is a simulation tool that allows for cost-effective research and development on learning from feedback, making research on instruction following and alignment more accessible.

\textbf{h2oGPT}~\cite{h2ogpt} is a framework for fine-tuning large language models and chatbot UI with document(s) question-answer capabilities. \textbf{H2O LLM Studio}~\cite{h2ollmstudio} is a no-code GUI large language model fine-tuning framework.

\textbf{PEFT}~\cite{peft} is a finetuning library developed by Huggingface that allows one to finetune large language models (LLMs) without having to train all of the model's parameters. It implements a variety of techniques including LoRA, prefix tuning, and p-tuning.

\textbf{MetaGPT}~\cite{metagpt} is capable of generating user stories, competitive analysis, requirements, data structures, APIs, documents, and more based on a single-line requirement. Within its system, MetaGPT encompasses product managers, architects, project managers, and engineers, effectively simulating the entire software company process. It also incorporates meticulously coordinated standard operating procedures (SOPs).




\section{Benchmark Evaluations}
To assess the performance of different language models across various tasks, we try to conduct an evaluation meticulously across multiple dimensions. The selected tasks scrutinize the model's general knowledge ability, information extraction capability, comprehension ability, mathematical skills, and competence in various academic disciplines.  The brief overview of these datasets is listed in table \ref{tab:5_text_dataset}. The chosen settings include zero-shot and few-shot scenarios, which are the two most important settings to evaluate the LLM's ability. The specific experiment setup and the results are discussed in detail in the following sections.

\begin{table*}[!ht]
\centering
\caption{Overview of benchmark datasets we evaluate for large language models.}
\label{tab:5_text_dataset}
\begin{tabular}{lccl}
\toprule
Corpora & Size & Latest updated time & Link \\
\midrule
BoolQ~\cite{clark2019boolq} &    15,492     &       2019      &   \url{https://github.com/google-research-datasets/boolean-questions}   \\
Hellaswag~\cite{zellers2019hellaswag} &	$\sim$70k&	2019&	\url{https://allenai.org/data/hellaswag}\\
WinoGrande~\cite{sakaguchi2021winogrande} &	$\sim$44k&	2019&	\url{https://winogrande.allenai.org/}\\
PIQA~\cite{Bisk2020}&	$\sim$21k&	2020&	\url{https://yonatanbisk.com/piqa/}\\
ARC~\cite{Clark2018ThinkYH}&	7,787	&2018&	\url{https://allenai.org/data/arc}\\
OpenbookQA~\cite{OpenBookQA2018}&	5,957	&2018&	\url{https://allenai.org/data/open-book-qa}\\
RACE~\cite{lai2017large}&	$\sim$100k&	2017&	\url{https://www.cs.cmu.edu/~glai1/data/race/}\\
DROP~\cite{Dua2019DROPAR}&	$\sim$96k	&2019&	\url{https://allenai.org/data/drop}\\
GSM8K~\cite{cobbe2021gsm8k}&	8,500	&2021&	\url{https://github.com/openai/grade-school-math}\\
TriviaQA ~\cite{JoshiTriviaQA2017} & $\sim$95k & 2017 & \url{http://nlp.cs.washington.edu/triviaqa/}\\
MMLU~\cite{hendryckstest2021}&	15,908	&2021&	\url{https://github.com/hendrycks/test}\\

\bottomrule
\end{tabular}
\end{table*}

\subsection{Evaluation Methods}
A zero-shot evaluation involves generating text for a specific, unique prompt that the model has never encountered before. The model needs to use its learned patterns and structures from the vast amount of training data to respond appropriately to this new prompt. Zero-shot capabilities are important for large language models, as they demonstrate their potential to generalize from learned data and apply knowledge in versatile scenarios.

Few-shot setting implies that these models can perform tasks by learning from a small number of examples as part of the prompt provided at inference time, even if those tasks were not explicitly trained on. Few-shot learning leverages the internalized knowledge of the model gained during training to adapt quickly to new tasks and underscores the generalization capabilities of large language models, which provides a practical and flexible way of using these models, as it enables them to perform a wide variety of tasks without requiring a massive amount of training data that is specific to each task.

\subsection{Zero-shot Evaluation}

\subsubsection{Dataset Summary}

For zero shot setting, we evaluate the models on the following datasets: BoolQ~\cite{clark2019boolq}, Hellaswag~\cite{zellers2019hellaswag}, WinoGrande~\cite{sakaguchi2021winogrande}, PIQA~\cite{Bisk2020}, ARC~\cite{Clark2018ThinkYH}, OpenbookQA~\cite{OpenBookQA2018}, RACE~\cite{lai2017large}, DROP~\cite{Dua2019DROPAR}, GSM8K~\cite{cobbe2021gsm8k}.

\textbf{BoolQ} \cite{clark2019boolq}. The BoolQ dataset, short for Boolean Questions, is a question-answering dataset for machine learning. It consists of 15942 examples, each containing a short passage, a yes/no question about the passage, and the correct answer. This dataset is used to train and evaluate models on the task of answering yes/no questions with some level of comprehension of the context provided in the passage.

\textbf{Hellaswag} \cite{zellers2019hellaswag}. HellaSwag stands for "Human-like Evaluation of Linguistic Ability and Simple Wordplay Generation", and the dataset contains a large number of multiple-choice questions. Each question presents a short context (often a description of a video) and four possible completions. The goal of the model is to predict the most plausible continuation.

\textbf{WinoGrande} \cite{sakaguchi2021winogrande}. Winogrande contains pairs of sentences that differ only in one or two words and that flip the correct reference of an ambiguous pronoun. The task for a machine learning model is to disambiguate the reference of the pronoun, which often requires an understanding of world knowledge, the context, and the likely intentions of the sentence's author.

\textbf{PIQA} \cite{Bisk2020}.  "Physical Interaction: Question Answering" (PIQA) is a task in natural language processing that focuses on the understanding of physical commonsense reasoning. The goal of this task would be to develop models that can answer questions about physical interactions in the real world, a fundamental aspect of human-like commonsense reasoning. This requires not only language understanding but also a basic knowledge of how the physical world works.

\textbf{ARC} \cite{Clark2018ThinkYH}. The ARC (AI2 Reasoning Challenge) dataset serves as a yardstick to assess the competence of AI systems in tackling intricate question-answering tasks. Unlike many existing QA datasets, the questions in the ARC dataset demand a higher level of reasoning skills. These might encompass abilities such as comprehending the physical world, grasping biological processes, or decoding cause-and-effect relationships. The dataset is divided into two categories: ARC Easy, comprising questions successfully answered by current AI systems, and ARC Challenge, which includes those that have stumped existing AIs.

\textbf{OpenbookQA} \cite{OpenBookQA2018}. The OpenBookQA dataset is an open-domain question answering dataset, which is designed to test a machine learning model's ability to answer questions that require external knowledge not present in the question itself. The premise of OpenBookQA is to provide a "book" of 1,326 core science facts and then ask 5,957 multiple-choice questions based on those facts. Each question can be answered by understanding and applying one or more of the scientific facts. However, most questions require additional broad, common knowledge to correctly answer.

\textbf{RACE} \cite{lai2017large}. The RACE (ReAding Comprehension from English Examinations) dataset is a large-scale reading comprehension dataset that was collected from English examinations in China designed for students in grades 3 through 12. RACE consists of nearly 100,000 questions that correspond to over 28,000 passages. One distinctive aspect of the RACE dataset is that it covers a broad range of topics and genres, reflecting the wide variety of material included in the English exams. This breadth makes it a valuable resource for testing the generalizability of machine learning models.

\textbf{DROP} \cite{Dua2019DROPAR}.  The Discrete Reasoning Over the content of Paragraphs (DROP) dataset is a machine reading comprehension dataset designed to assess the ability of an AI model to process complex natural language questions and perform reasoning over paragraphs of text. The questions in the DROP dataset often require complex reasoning skills, such as performing arithmetic operations, comparing entities, or understanding the sequence of events. They are generally formulated in a way that the AI model has to comprehend multiple parts of the passage to be able to answer correctly.

\textbf{GSM8K} \cite{cobbe2021gsm8k}. The GSM8K dataset is a collection of 8,500 math word problems, carefully designed for grade school students. These problems, crafted by expert problem writers, exhibit a wide range of linguistic diversity. The unique aspect of these problems is that they are multi-step in nature, necessitating anywhere from 2 to 8 separate calculations to arrive at the final solution. The problems are calibrated to challenge but not overwhelm middle school students, thereby making the dataset a useful resource for developing and testing algorithms for multi-step mathematical reasoning.

\subsubsection{Results}
The performance results derived from the datasets under discussion can be conveniently found and reviewed in Table \ref{tab:zero-shot-text}. This table provides a comprehensive understanding of different models' performance in the zero-shot setting, showcasing their strengths and weaknesses in diverse tasks. These models include LLaMA2-7B, Stanford Alpaca-7B, Vicuna(FastChat)-7B, ChatGLM-6B, Alpaca-LoRA et cetera. Other papers' results may differ from the results presented herein, primarily due to variations in the prompts employed. Nevertheless, by comparing the outcomes with the LLaMA-7B baseline, we can still obtain a general understanding of the model's performance. The table reveals several intriguing observations. 

The performance disparities among the language models can primarily be attributed to the tuning methods employed during their development. It is worth noting that while many of the models analyzed in this study are based on the LLaMA-7B architecture, their individual performances vary significantly. Some models exhibit a strong average performance, while others fall short in comparison. This highlights the crucial role played by tuning techniques in shaping the models' overall effectiveness. 

Moreover, the study underscores the varying efficacy of language models across different tasks. It becomes evident that these models excel in certain areas while displaying weaknesses in others. No single model can claim complete dominance over all datasets and tasks. However, when considering the overall performance, the models exhibit distinct and diverse abilities relative to one another. 

Lastly, the language model demonstrates a favorable performance in tasks involving questions with options, while its effectiveness diminishes considerably in generative tasks. This discrepancy can be easily comprehended as the act of generating content is typically more challenging than performing classification tasks. Generating coherent and contextually appropriate responses requires a deeper understanding of language and context, making it a more complex endeavor compared to the relatively straightforward classification of predefined options.

\begin{table*}[t]  
\centering  
\caption{Results (accuracy \%) on general QA datasets including BoolQ \cite{clark2019boolq}, Hellaswag\cite{zellers2019hellaswag}, WinoGrande\cite{sakaguchi2021winogrande}, PIQA\cite{Bisk2020}, ARC\cite{Clark2018ThinkYH}, OpenbookQA\cite{OpenBookQA2018}, RACE\cite{lai2017large}, DROP\cite{Dua2019DROPAR} and GSM8K\cite{cobbe2021gsm8k} for different models under zero-shot setting. \textcolor{orange!80!red}{Orange} indicates best performance, while \textcolor[HTML]{FEB8DD}{pink} indicates the second and third best.}  
\resizebox{\textwidth}{!}{
\begin{tabular}{l|c|c|c|c|c|c|c|c|c|c|c}  
\toprule
\multirow{2}{*}{Method} & \multirow{2}{*}{BoolQ} & \multirow{2}{*}{Hellaswag} & \multirow{2}{*}{WinoGrande} & \multirow{2}{*}{PIQA} & \multicolumn{2}{c|}{ARC} & \multirow{2}{*}{OBQA} & \multirow{2}{*}{TriviaQA} & \multirow{2}{*}{RACE} & \multirow{2}{*}{DROP} & \multirow{2}{*}{GSM8K} \\
& & & & & easy & challenge & & & & & \\
\midrule
LLaMA2-7B & \cellcolor[HTML]{FEE8DD} 75.72 & 72.86 & \cellcolor[HTML]{FEE8DD} 68.12 & 78.24 & 68.28 & 38.56 & \cellcolor[HTML]{FC8D59} 47.8 &  \cellcolor[HTML]{FC8D59} 64.53 & \cellcolor[HTML]{FEE8DD} 41.8 & 12.65 & \cellcolor[HTML]{FEE8DD} 6.82 \\
Vicuna(FastChat)-7B \cite{vicuna} & \cellcolor[HTML]{FC8D59} 76.57 & 70.66 & 67.25 & 77.80 & 65.07 & 39.68 & 40.8 & 58.95 & 41.24 & 12.18 & \cellcolor[HTML]{FEE8DD} 6.44 \\
LLaMA-7B & 73.15 & \cellcolor[HTML]{FEE8DD} 72.99 & 67.09 & \cellcolor[HTML]{FEE8DD} 78.35 & 67.34 & 38.23 & 42.4 & \cellcolor[HTML]{FEE8DD} 62.42 & 40.00 & 3.18 & 2.65 \\
Stanford Alpaca-7B \cite{alpaca} & \cellcolor[HTML]{FEE8DD} 74.71 & 72.64 & 67.88 & 78.29 & \cellcolor[HTML]{FEE8DD} 71.84 & \cellcolor[HTML]{FEE8DD} 42.41 & \cellcolor[HTML]{FEE8DD} 42.8 & \cellcolor[HTML]{FEE8DD} 61.7 & \cellcolor[HTML]{FC8D59} 43.64 & \cellcolor[HTML]{FEE8DD} 14.85 & 4.7 \\
Alpaca-LoRA \cite{alpaca-lora} &  74.34 & \cellcolor[HTML]{FEE8DD} 74.03 & \cellcolor[HTML]{FC8D59} 68.67 & \cellcolor[HTML]{FEE8DD} 78.99 & 68.66 & \cellcolor[HTML]{FEE8DD} 42.83 & 42.6 & 61.19 & 41.15 & 3.48 & \cellcolor[HTML]{FC8D59}9.97 \\
ChatGLM-6B \cite{du2022glm, zeng2023glm-130b}   & 71.18 & 62.12 & 67.09 & 76.35 & 62.21 & 32.08 & 40.8 & 56.62 & 40.07 & 5.52 & 2.15 \\
Databricks Dolly-v2-7b \cite{dolly} & 65.02 & 68.96 & 61.33 & 74.48 & 67.17 & 37.71 & 39.6 & 61.06 & 38.95 & 10.26 & 2.27 \\
StableLM-Tuned-Alpha-7B \cite{stablelm} & 62.54 & 53.64 & 54.78 & 71.76 & 52.40 & 31.06 & 33.4 & 58.96 & 33.88 & 10.28 & 3.34 \\
MPT-7B-Instruct \cite{mpt7b} & 72.60 & \cellcolor[HTML]{FC8D59} 77.10 & 67.56 & \cellcolor[HTML]{FC8D59} 80.41 & \cellcolor[HTML]{FC8D59} 76.47 & \cellcolor[HTML]{FC8D59}42.92 & \cellcolor[HTML]{FEE8DD} 43.6 & 24.69 & \cellcolor[HTML]{FEE8DD} 41.91 & \cellcolor[HTML]{FEE8DD} 15.95 & 3.18 \\
PandaLM-7B \cite{PandaLM} & 70.61 & 54.18 & \cellcolor[HTML]{FEE8DD} 67.96 & 78.18 & \cellcolor[HTML]{FEE8DD} 69.49 & 37.88 & 41.00 & 0.00 & 39.90 & 13.97 & 3.64 \\
MOSS \cite{moss} & 65.81 & 60.11 & 60.38 & 72.52 & 64.48 & 34.73 & 38.0 & 10.99 & 33.21 & 7.03 & 3.02 \\
Open-Assistant \cite{openassistant} & 62.63 & 55.08 & 56.20 & 69.75 & 55.85 & 30.12 & 33.40 & 5.33 & 34.07 & 13.77 & 5.46 \\

Pythia-Chat-Base-7B \cite{biderman2023pythia}  & 52.39 & 68.31 & 62.59 & 75.73 & 68.22 & 35.32 & 40.4 & 62.1 & 34.93 & 5.18 & 1.29 \\
BELLE-7B \cite{belle} & 42.51 & 55.73 & 64.21 & 71.33 & 55.09 & 31.06 & 32.6 & 3.01 & 40.19 & \cellcolor[HTML]{FC8D59} 16.12 & 3.28 \\

\bottomrule
\end{tabular}  
}
\vspace{2mm}
\label{tab:zero-shot-text}  
\end{table*}



\subsection{Few-shot evaluation}

\subsubsection{Dataset Summary}

In the context of a few-shot learning setup, we undertake the evaluation of the models using three distinct datasets: MMLU\cite{hendryckstest2021} and TriviaQA\cite{JoshiTriviaQA2017}. Our experimental design incorporates a varied number of shots, including the 0-shot setting, which serves as a baseline, followed by the 1-shot, 3-shot, and 5-shot respectively.

\textbf{MMLU} \cite{hendryckstest2021}. The Massive Multitask Language Understanding (MMLU) dataset is a versatile tool designed to gauge the knowledge that models gain during pretraining. The dataset encompasses a wide array of 57 subjects, spanning areas such as STEM, humanities, social sciences, and more. These subjects vary from elementary to advanced professional levels of difficulty. MMLU does not only test world knowledge but also problem-solving abilities, covering traditional fields like mathematics and history as well as specialized ones such as law and ethics. 

\textbf{TriviaQA} \cite{JoshiTriviaQA2017}. TriviaQA is a reading comprehension dataset containing over 650K question-answer-evidence triples. It's specifically designed for the training and evaluation of reading comprehension systems, especially ones focused on multi-sentence and multi-hop reasoning. The questions in TriviaQA are divided into two main categories: web domain and Wikipedia domain. TriviaQA questions often require complex reasoning, and the associated evidence documents often contain more than one sentence that can help answer the question, which adds to the complexity of tasks built on this dataset. This complexity and the scale of the dataset make TriviaQA a common choice for developing and evaluating advanced natural language processing and question answering models.

\subsubsection{Results}

The evaluation of various models' performance on the MMLU and TriviaQA datasets can be accessed and in Table \ref{tab:few-shot-text}. This table provides an assessment of the models' capability in few-shot learning, a setting where they learn from a small number of examples. Upon comparing the performance metrics, it becomes clear that these models exhibit differences not only in their performance in distinct tasks but also in their few-shot learning capabilities. Some models, such as LLaMA, demonstrate a positive correlation between their performance and the number of examples provided; their performance improves as the quantity of data increases. However, other models, like Stanford's Alpaca-7B and Vicuna(FastChat)-7B, do not exhibit this propensity. This comparative analysis in Table \ref{tab:few-shot-text} thus sheds light on the unique learning behaviors of these models, offering valuable insights for potential improvements and future research.

From the table, we can draw several noteworthy observations. Firstly, it is evident that the performance of the language model tested here does not consistently improve with an increase in the number of few-shot examples. This lack of performance enhancement can potentially be attributed to the model's relatively small size and its limited ability to effectively utilize few-shot learning techniques. Consequently, the model struggles to glean meaningful insights from the provided examples.

Secondly, the performance of the language model remains relatively stable across different few-shot settings. Despite exhibiting diverse tendencies, certain settings display a positive correlation between performance and the number of shots, while others demonstrate the opposite trend. Nonetheless, the overall difference in performance between each setting is relatively minor. Thus, if a model excels under the zero-shot setting, it is likely to perform well on average as well.

Thirdly, it is crucial to acknowledge that some of the tested language models exhibit subpar performance. These models may require more appropriate prompts or additional fine-tuning steps to acquire the necessary knowledge and improve their overall performance. It is possible that these models are not effectively leveraging the available resources and require further optimization to enhance their abilities.

\begin{table*}[t]  
\centering  
\caption{Results (accuracy \%) on general QA datasets including MMLU\cite{hendryckstest2021} and TriviaQA\cite{JoshiTriviaQA2017} for different models under few-shot setting. \textcolor{orange!80!red}{Orange} indicates best performance, while \textcolor[HTML]{FEB8DD}{pink} indicates the second and third best.}  
\resizebox{!}{!}{
\begin{tabular}{l|cccc|cccc}  
\toprule
\multirow{2}{*}{Method} & \multicolumn{4}{c|}{MMLU} & \multicolumn{4}{c}{TriviaQA} \\
& 0 shot & 1 shot & 3 shot & 5 shot & 0 shot & 1 shot & 3 shot & 5 shot \\
    \midrule
LLaMA2-7B & 42.7 & 43.1 & 43.2 & 43.8 & 64.53 & \cellcolor[HTML]{FEE8DD} 65.72 & 65.68 & \cellcolor[HTML]{FC8D59} 66.11 \\
Vicuna(FastChat)-7B \cite{vicuna} & \cellcolor[HTML]{FC8D59}45.5 & 44.2 & \cellcolor[HTML]{FEE8DD} 44.3 & \cellcolor[HTML]{FEE8DD} 44.9 & 58.95 & 63.16 & 64.24 & 64.83 \\
Stanford Alpaca-7B \cite{alpaca} & 40.9 & 39.1 & 39.7 & 40.4 & 61.7 & 63.55 & 64.29 & 64.32 \\
Alpaca-LoRA \cite{alpaca-lora} & 38.0 & 34.0 & 35.3 & 36.0 & 61.19 & 63.5 & 64.23 & 64.45 \\
LLaMA-7B & 28.5 & 32.9 & 34.1 & 35.2 & 62.42 & 64.63 &  65.58 & \cellcolor[HTML]{FEE8DD} 65.95 \\
ChatGLM-6B \cite{du2022glm, zeng2023glm-130b} & 35.2 & 35.5 & 36.0 & 36.5 & 56.62 & 57.79 & 58.92 & 59.08 \\
Databricks Dolly-v2-7b \cite{dolly} & 25.43 & 24.86 & 25.16 & 25.22 & 61.06 & 62.58 & 62.49 & 62.79 \\
StableLM-Tuned-Alpha-7B \cite{stablelm} & 25.29 & 25.32 & 26.17 & 24.45 & 58.96 & 59.76 & 60.01 & 60.74 \\
Pythia-Chat-Base-7B \cite{biderman2023pythia}   & 28.36 & 28.21 & 28.76 & 29.12 & 62.1 & 63.17 & 63.09 & 63.36 \\
MPT-7B-Instruct \cite{mpt7b} & 35.63 & 33.28 & 35.42 & 34.71 & 24.69 & 35 & 36.81 & 36.99 \\
BELLE-7B \cite{belle} & 40.36 & 34.03 & 36.21 & 35.19 & 3.01 & 4.07 & 5.3 & 5.32 \\
MOSS \cite{moss} & 30.93 & 34.51 & 34.88 & 32.86 & 10.99 & 13.98 & 3.57 & 0.72 \\
Open-Assistant \cite{openassistant} & 28.34 & 30.22 & 29.56 & 30.55 & 5.33 & 6.43 & 7.43 & 8.11 \\
PandaLM-7B \cite{PandaLM} & 30.3 & 27.7 & 29.0 & 28.0 & 0.00 & 0 & 0 & 0 \\
\bottomrule
\end{tabular}  
}
\vspace{2mm}
\label{tab:few-shot-text}  
\end{table*}

\section{Human Evaluation}

With the emergence of more astonishing large language models (LLMs), evaluation benchmarks fall short of fully assessing the capabilities of these LLMs. Additionally, existing automatic reference-based metrics have displayed a pronounced inconsistency with human evaluation~\cite{goyal2022news, zhang2023benchmarking}. Therefore, we augment our analysis with comprehensive human evaluation experiments to assess the model for human alignment in this section.

\subsection{Evaluation Setup}

\begin{figure*}[!ht]  
\centering  
\includegraphics[width=1.0\linewidth]{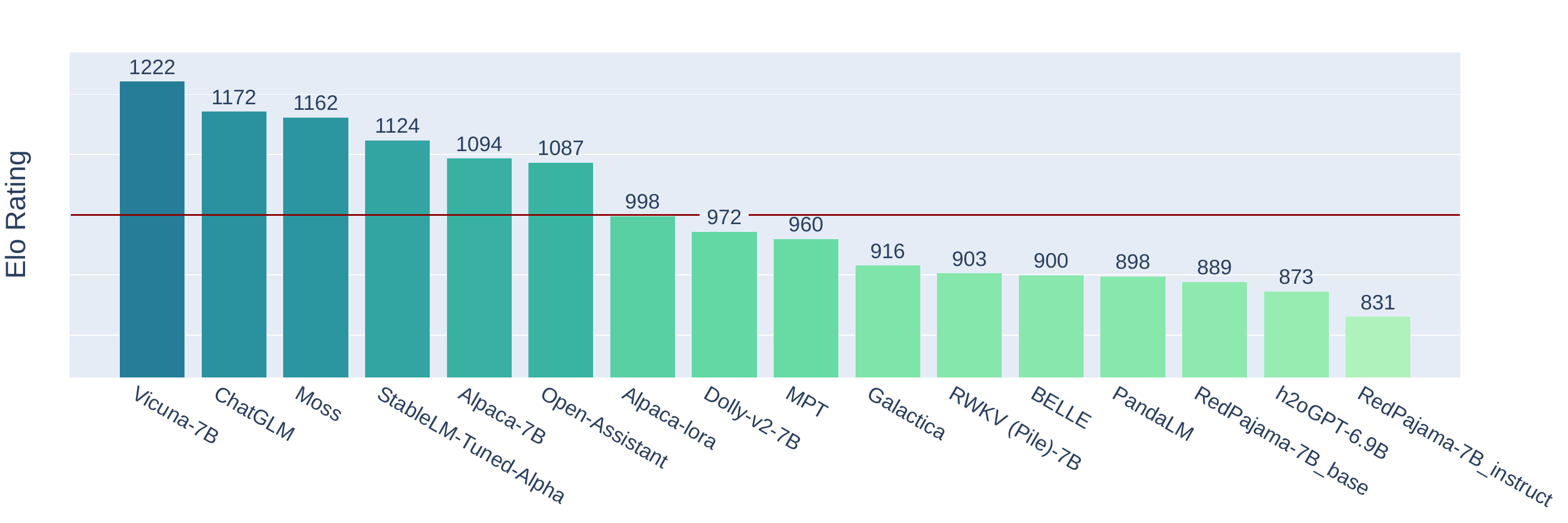}    
\caption{Elo rating leaderboard after $1000$ rounds of randomized matches. The initial score of $1000$ is represented by the red line.}
\label{fig:elo_ranking}
\vskip -0.1in
\end{figure*}

Benchmarking models through human evaluation presents significant challenges, including the open-endedness of the questions and the subjectivity of human evaluators. Moreover, the need to evaluate many models (16 models in total\footnote{We do not include the LLaMA2-Chat and LLaMA2-based fine-tuning models in our human evaluation since they are recently released after our human evaluation is done. However, from the competitive results in benchmark evaluations, we believe LLaMA2-Chat and LLaMA2-based tuning models (such as Vicuna-1.5) can stay in top rank as strong models.}) complicates the objective of accurately scoring the models. To address this, we employ a pairwise comparison approach to evaluate model performance, as it's relatively easier to determine a winner between two options. We design 50 diverse questions and used them to evaluate 16 models with approximately 7 billion parameters each (models can be found in~\ref{fig:elo_ranking}). Subsequently, we apply the Elo rating system to score the final results~\cite{vicuna, dettmers2023qlora, zheng2023judging}.

\paratitle{Question design.} 
Informed by numerous human evaluation prompts~\cite{vicuna, selfinstruct, ganguli2022red}, we meticulously craft 50 high-quality questions spanning 9 common categories: generic, writing, reasoning, coding, math, physics, chemistry, biology, and toxic. We carefully design 5 to 10 questions for each category. Several sample questions are listed in Table~\ref{tab:6_sample_questions}. For categories like generic and writing, our goal is to assess the models' consistency, fluency, and written capabilities. In contrast, for categories like reasoning and coding, our aim is to evaluate the quality of the models' chains-of-thought~\cite{wei2022chain} ability. For questions pertaining to math, physics, chemistry, and biology, we seek to assess the models' grasp of essential knowledge, such as atomic mass, as well as their ability to apply this knowledge to solve problems. Concurrently, we also want to investigate the hallucination issues of models. In the category of toxic identification, our objective was to ensure the model maintains a positive, fair, and unbiased output.

\begin{table*}[!ht]
\footnotesize
\caption{Sample questions for human evaluation. }
\label{tab:6_sample_questions}
\centering
\resizebox{!}{!}{
\begin{tabular}{@{}p{0.10\linewidth}|p{0.87\linewidth}@{}}
\toprule
Category  & \multicolumn{1}{c}{Sample Questions} \\ 
\midrule
Generic & Provide me a detailed 5-day travel itinerary in Beijing. My budget is 10,000 RMB. \\
\midrule
Writing  &  An Area Chair (AC) is to decide whether a paper should be accepted or not based on the reviewer's comments. If you are an AC and you want to accept the paper, but four of your reviewers give negative comments, how to write a paper decision to argue for accepting this paper. \\ 
\midrule
Math  &  Given five points A(2, 4), B(-1, 2), C(-3, -3), D(2, -4), and E(4, -2), please calculate the area of the pentagon formed by these points. \\
\midrule
Physics & Calculate the energy released when a proton and a neutron combine to form a deuterium nucleus. \\
\midrule
Chemistry & Balance the following chemical equation: HClO$_3$ = HClO$_4$ + Cl$_2$↑ + O$_2$↑ + H$_2$O. \\
\bottomrule
\end{tabular}
}
\end{table*}

\paratitle{Elo Rating.} The Elo rating system~\cite{elo1978rating} is a mathematical model used in competitive games to calculate players' relative performance, adjusting ratings based on performance against other players. For each pair comparison, if the outcome was a tie, the scores of the two models remained unchanged. If there was a winner, the winning model's score was increased, while the losing model's score was decreased. The magnitude of these changes was proportional to the difference in scores between the two models at the start of the comparison, ensuring that an unexpected win by a lower-ranked model led to a larger adjustment than a predictable win by a higher-ranked model.

It should be noted that the Elo rating scores differ with the comparison order. To mitigate this variability, we conducted multiple rounds of calculation. After each round of computation, the order is randomly shuffled. This procedure was performed for a total of $1000$ iterations, aiming to enhance the stability and robustness of the final rating outcomes.

\subsection{Evaluation Process.}
For the $16$ models under consideration, each question generates $C^2_{16}=120$ pairwise comparisons, yielding a total of $6000$ pairs across all $50$ questions. For each pair of answers, evaluators should judge whether ``model A wins'', ``model B wins'', or if it's ``a tie''.

Besides, to validate human consistency, we first proportionately selected $20$ questions from the nine categories, resulting in $2400$ question-answer pairs. These pairs were split amongst two groups of annotators, with two individuals per group, and each individual responsible for $1200$ pairs. The aim was to gauge the level of agreement between the groups, hoping for a high degree of consistency in their responses.
Following this, the remaining $3600$ pairs were distributed among nine additional annotators, with each assigned a random selection of $400$ pairs.

When counting the final scores, the two sets of $2400$ responses, used for human consistency, were consolidated. (1) If the responses from the two annotators aligned, their shared view was taken as the result. (2) If one annotator indicated a tie, the judgment of the other was taken as the answer. 
 (3) If one annotator believed that 'model A wins' and the other that 'model B wins,' the models were considered tied. After this simplification process, these results were concatenated with the $3600$ pairs' responses, resulting in a full $6000$ response for each pair.

\subsection{Results and analysis}
\noindent \textbf{Elo Rating Results.} Figure~\ref{fig:elo_ranking} illustrates the diverse range of Elo ratings. The initial Elo rating for all models was set to $1000$, and a K-factor of $16$ was utilized to control the maximum change in rating. Among the $16$ models, Vicuna-7B ranks at the top, exhibiting the highest Elo rating of $1222$. The second and third positions are closely contested by ChatGLM and Moss. Moreover, the majority of models, ranging from 7th to 15th place, demonstrate closely aligned performance levels. From another perspective, the Elo rating system does indeed exhibit discriminatory power, indicating a distinct performance hierarchy among the models.


\begin{figure*}[!ht]
\centering
\begin{subfigure}{0.5\textwidth}  
  \centering  
  \includegraphics[width=\linewidth]{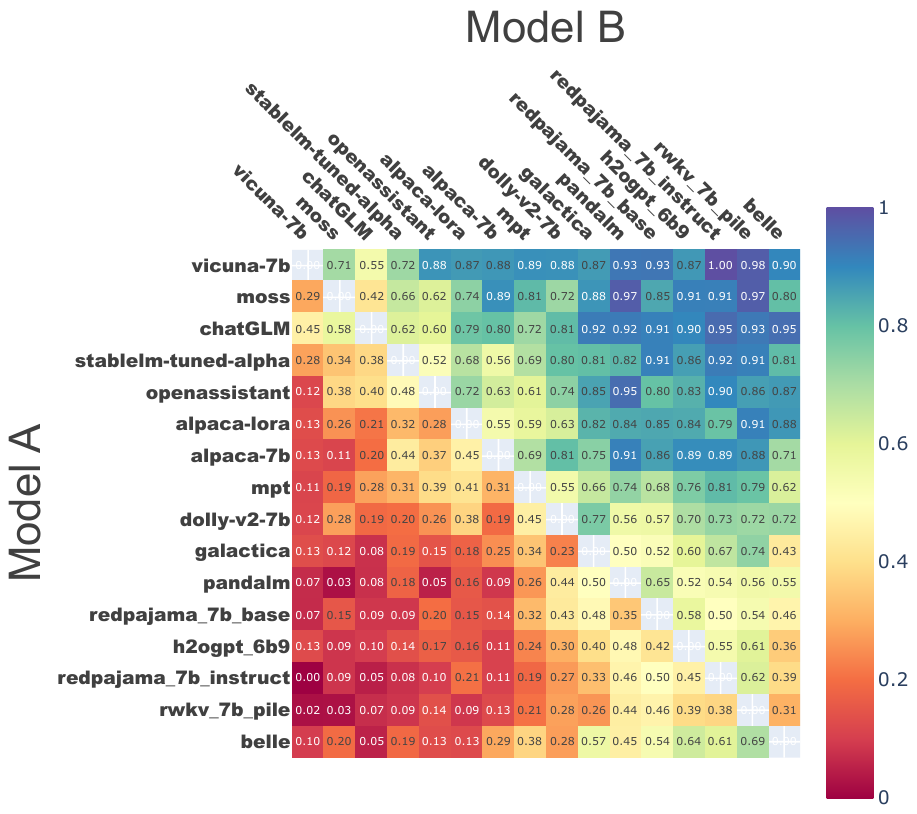}    
  \caption{Fraction of Model A wins for all non-tied battles.}
  \label{fig:win-rate-real}
\end{subfigure}
\vskip 0.3in  
\begin{subfigure}{0.5\textwidth}  
  \centering  
  \includegraphics[width=\linewidth]{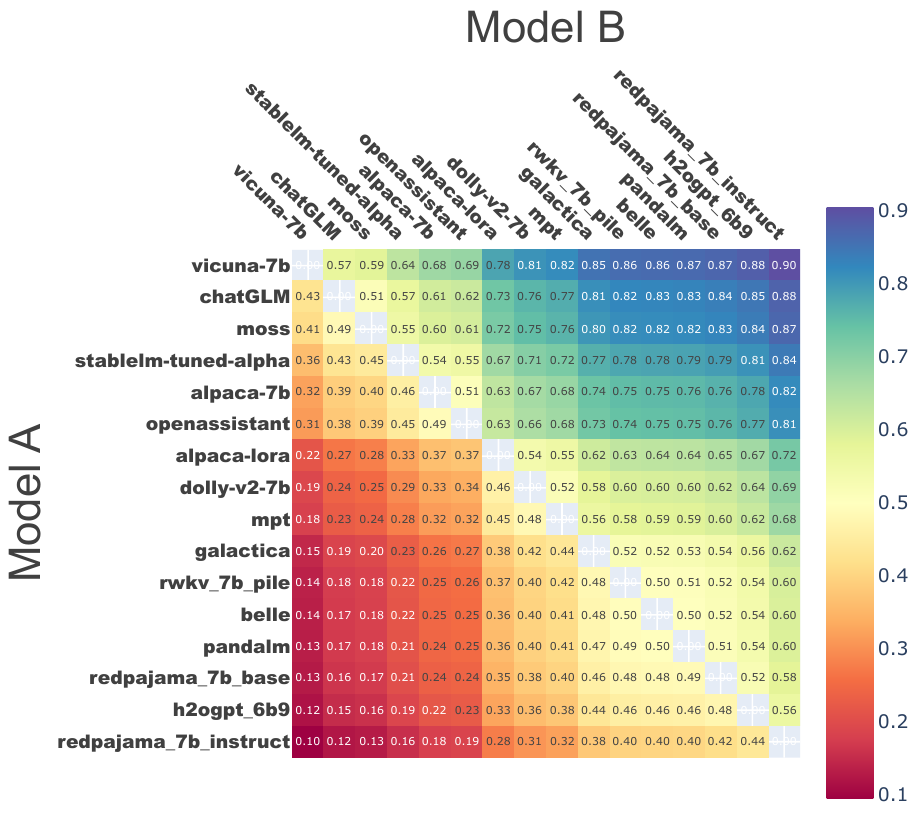}    
  \caption{Predicted win rates using Elo rating results.}
  \label{fig:win-rate-pred}
\end{subfigure}
\caption{Pairwise win-rate visualization for Model A in an A vs. B battle. The Elo ratings can predict win rates relatively well.}
\label{fig:vis-win-rate}
\vskip -0.1in
\end{figure*}

\begin{figure*}[!ht]  
\centering  
\includegraphics[width=0.9\linewidth]{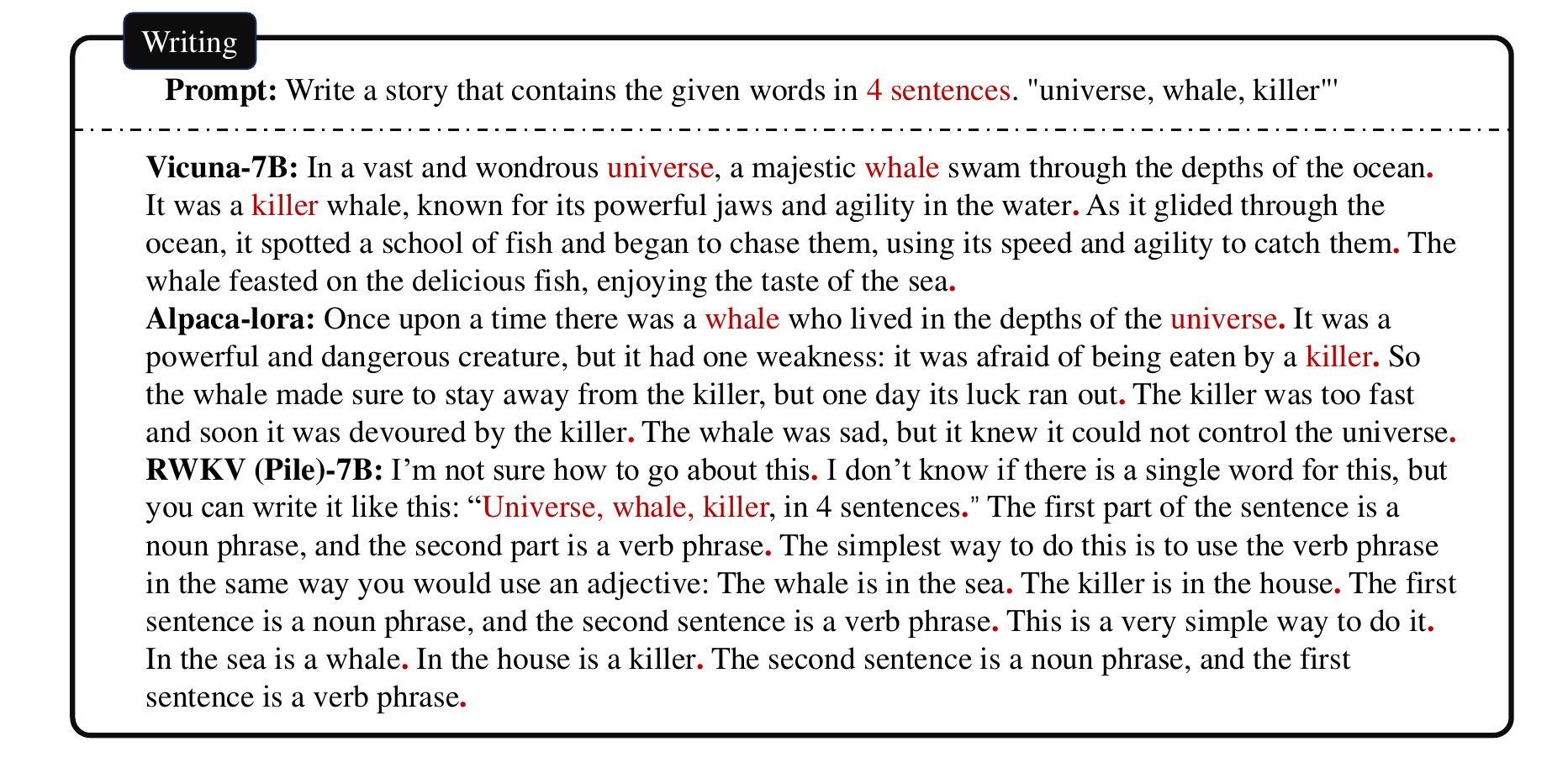}    
\caption{Comparative analysis of responses by top-ranked Vicuna-7B, seventh-ranked Alpaca-lora, and eleventh-ranked RWKV (Pile)-7B models to a question in the ``Writing'' category.}
\label{fig:human-eval-case}
\vskip -0.1in
\end{figure*}

\paratitle{Win-rate.}
The Elo rating score also serves as a match outcome predictor. Within a certain range, each $10$-point difference in Elo is approximately a difference of $1.5$\% in win-rate. Consequently, we are able to construct a pairwise win-rate heatmap, based on Elo rating predictions, as displayed in Figure~\ref{fig:win-rate-pred}.
Furthermore, we draw a separate heatmap representing the actual win-rates between models in Figure~\ref{fig:win-rate-real} for comparison. These illustrations suggest that the Elo rating system effectively captures the win-rates. For instance, when considering Vicuna-7B and ChatGLM, the $50$-point Elo score difference predicts a win-rate of $57$\% for Vicuna-7B against ChatGLM. This estimation closely aligns with the actual win-rate of $55$\%.

\paratitle{Human consistency.} We evaluate the consistency between annotators using a tie-discounted accuracy~\cite{zhou2023lima}. In this measurement, a full point is awarded when both annotators concur, half a point is given if only one annotator denotes a tie, and no points are awarded in other situations. After gauging this agreement across the 20 selected questions, an agreement score of 80.02 out of 100 between two evaluators in two groups is observed. Hence, despite the inherent subjectivity of different human annotators, there is an acceptable level of concordance.

\paratitle{Cases.} In our case study, we examine the responses of models to a ``Writing'' question. In this task, models were asked to compose a story containing three given keywords within a sentence limit. The top-ranked Vicuna-7B effectively generated a coherent story using four sentences. The seventh-ranked Alpaca-lora managed to meet the requirements with coherent sentences, but it included an extra sentence and some factual errors, such as ``killer eating whale.'' On the contrary, the eleventh-ranked RWKV (Pile)-7B produced an excessive amount of content and failed to comprehend the question's requirements, resulting in a rather disorganized output.  These findings align with their respective ranks.

\paratitle{Limitations.}
While our study provides valuable insights, it has limitations worth noting. Due to resource constraints, our $50$-question set is not fully representative, overlooking certain model capabilities like multi-linguality. Moreover, our computing-related questions might have been too challenging for medium-sized models. Additionally, the Elo rating system is unstable, as it produces inconsistent rankings even when repeated with different orders.


\section{Multimodal GPT Models and Evaluation } 

Although LLMs have demonstrated their effectiveness in natural language processing, their ability to comprehend other modalities, such as vision and audio, remains a challenge.
Table~\ref{tab:7_mmllm_summary} presents the comparison of recent multi-modal large language models on the number of parameters and their architectures. 

\subsection{History of Multimodal Foundation Models}
We first recap the history of the multimodal foundation models. Without available large language models, previous vision-language models usually utilize BERT~\cite{devlin2018bert} or RoBERTa~\cite{liu2019roberta} as the text encoder to extract textual features.
Due to the limitations in model size, these models have not demonstrated amazing reasoning abilities and can only solve simple visual question answering tasks~\cite{antol2015vqa}.
Recently, with the emergence of large language models~\cite{touvron2023llama} and their demonstrated powerful reasoning abilities, several vision-language multimodal works have introduced large language models to accomplish complex multimodal understanding tasks.
Flamingo~\cite{alayrac2022flamingo} first proposed to combine the powerful pretrained vision-only and language-only models for multimodal understanding.
BLIP-2~\cite{li2023blip} utilizes pre-trained FlanT5 models in a frozen state and employs a Q-Former for extracting visual features to be used as input for the LLMs. 
MiniGPT4~\cite{zhu2023minigpt} leverages the same pre-trained visual encoder and Q-Former as BLIP-2 but adopts Vicuna as the LLM. Furthermore, MiniGPT4 extends the training process by utilizing longer image captions generated by ChatGPT.
LLaVA~\cite{liu2023visual} takes a direct approach by projecting the output of a visual encoder as input to a large language model.
In contrast, mPLUG-owl~\cite{ye2023mplug} incorporates low-rank adaptation to finetune an LLaMA model. This finetuning process involves utilizing both text instruction data and vision-language instruction data obtained from LLaVA.
VPGTrans~\cite{zhang2023transfer} is a two-stage transfer framework that can effectively speed up the transfer learning process among LLMs without compromising performance.
Otter~\cite{li2023otter} is a multimodal model trained on MIMIC-IT with a more powerful instruction-following ability and better performance in in-context learning.
MultiModal-GPT~\cite{gong2023multimodal} is built upon the OpenFlamingo model and finetuned with Low-rank Adapter.

\subsection{Models Summary}
For the multimodal GPT, we focus on the following aspects:
\begin{enumerate}
    \item \textbf{Model Structure}: Multimodal GPTs usually consist of a vision encoder, a vision-to-language converter, and a large language model. The vision encoder aims to extract the visual information from the image, which is usually a vision transformer initialized by vision-language pre-training (e.g., CLIP and Flamingo). The vision-to-language converter, which projects the visual embedding into the language embedding space, is designed to minimize the modal gap between vision and language. The large language model is finally integrated with a converter to generate desired text sentences, e.g., answers to visual questions. 
    \item \textbf{Training Dataset}: As is well known, the quality of the dataset largely determines the performance of a model. Therefore, constructing a suitable pretraining dataset for multimodal language models has become a major focus of research. Traditional multimodal pretraining datasets mainly consist of image-text pairs, where each image is associated with a corresponding sentence or caption, such as CC~\cite{sharma2018conceptual}, VG~\cite{krishna2017visual}, SBU~\cite{ordonez2011im2text}, COCO~\cite{lin2014microsoft}, and others. However, models trained on such single-turn data have not shown satisfactory performance in the context of dialogue. To obtain multi-turn dialogue data, some approaches have been proposed using large language models such as ChatGPT and GPT-4. These models generate conversations or detailed descriptions based on image prompts or object information. This allows for the creation of dialogue-based datasets and expands the capabilities of multimodal language models in the domain of conversational tasks.
    \item \textbf{Pre-text Tasks}: Due to the utilization of the Language-Image Model (LIM) as the interface for model interaction, the training task of a multimodal LLM primarily focuses on the next token prediction. This task requires the model to comprehend complex multimodal content based on input images or textual information and generate appropriate responses.
    \item \textbf{Training Strategy}: Due to the enormous scale of large language models, training the entire set of parameters can be extremely resource-intensive. Therefore, some methods have proposed training only a subset of the model's parameters using techniques like Lora or Prompt Tuning, which ensures successful model training while significantly reducing the training cost. 
    Both Lora and Prompt Tuning provide strategies to mitigate the computational burden associated with training large language models in multimodal foundation models, making it feasible to train them in a more resource-efficient manner.
\end{enumerate}

Next, we will present the details of several multi-modal GPT models, i.e., MiniGPT4 \cite{zhu2023minigpt}, LLaVA \cite{liu2023visual}, VPGTrans \cite{zhang2023transfer}, mPLUG-Owl \cite{ye2023mplug}, Otter \cite{li2023otter} and MultiModal-GPT \cite{gong2023multimodal}.

\textbf{MiniGPT4}~\cite{zhu2023minigpt} consists of a vision encoder with a pre-trained ViT and Q-Former, a single linear projection layer, and a large language model Vicuna.
The model is trained on both image-text pair datasets and carefully curated a high-quality image-text dataset.

\textbf{LLaVA}~\cite{liu2023visual} consists of an image encoder, projection head, and a pre-trained large language model. The model is trained on a two-stage instruction-tuning procedure. 

\textbf{VPGTrans}~\cite{zhang2023transfer} aims to reduce the dispensable training costs of the visual prompt generator (VPG) by transferring pre-trained VPG. Specifically, they inherit the pre-training weight from relatively small models to large counterparts and fine-tune the VPG for fewer epochs.
\par
\textbf{mPLUG-Owl}~\cite{ye2023mplug} consists of a vision foundation model to encode the visual knowledge, a visual abstractor module to summarize visual information, and a large language model to generate the answer.
To evaluate the performance of the multimodal large language model, they proposed an instruction evaluation set OwlEval and employed manual evaluation metrics to rate the model's response.
\par
\textbf{Otter}~\cite{li2023otter}, a multi-modal model with in-context instruction tuning based on the OpenFlamingo model. 
They presented Multi-Modal In-Context Instruction Tuning (MIMIC-IT) dataset to further improve instruction comprehension capabilities.
Specifically, each sample in the MIMIC-IT dataset consists of a queried image-instruction-answer triplet and several context triplets that correlate to the queried triplet.
\par
\textbf{MultiModal-GPT}~\cite{gong2023multimodal} is also built upon the open-flamingo model which consists of a vision encoder CLIP, a perceiver resampler and a language decoder LlaMA.
The training datasets include language-only instruction-following data and vision-language instruction-following data.
Both of these datasets share the same unified instruction template during the training process.

\subsection{Benchmark Evaluations}
To conduct a representative benchmark evaluation and especially dig the knowledge of the multimodal language models, we evaluate these multimodal GPT models on a scientific question answering dataset (ScienceQA) instead of the commonly used visual question answering dataset, which is supposed to be a better testbed for different capability evaluation, such as knowledge and reasoning ability. 
Science Question Answering (ScienceQA) dataset~\cite{lu2022learn} is a vision-language question answering benchmark in the science domain.
ScienceQA contains about 21k multiple-choice questions, covering rich domains and diverse topics.
Moreover, most questions have a corresponding lecture and explanation, facilitating the training for the chain-of-thought.
Following the evaluation setting in ScienceQA~\cite{lu2022learn}, we conduct a 2-shot setup following the format of QCM$\to$A.
The task requires the model to generate the correct answer (A) from the open set when the question text (Q), context text (C), and multiple options (M).

Table~\ref{tab:7_sqa_benchmark} shows the results of the ScienceQA test dataset
Among the open-sourced large language models, Vicuna-13B-v1 achieved the best overall performance on the ScienceQA dataset which is even higher than some multimodal methods.
Alpaca-Lora obtains higher performance in language science problems while acquiring lower scores in nature science and social science problems.
For other multimodal counterparts, MiniGPT4-13B and VPGTrans, utilizing the more powerful large language model Vicuna, significantly outperformed the other multimodal methods, achieving about 50\% overall accuracy, indicating that larger vision encoder (e.g., ViT-G) and stronger large language model (e.g., Vicuna) can further improve the performance.

Table~\ref{tab:7_sqa_img_benchmark} presents the results on the subset data with image context.
Because of more parameters and a more powerful language model (vicuna) integrated, MiniGPT4-13B suppresses other methods and achieves 52.9\% overall accuracy.
For the model with 7B size, VPGTrans obtained the best results in all terms.
Specifically, MiniGPT4-13B achieves the best score in natural science and social science questions while VPGTrans obtains the highest accuracy in language science questions.
In terms of grades, MiniGPT4-13B performs better on low-grade problems, while VPGTrans performs better on high-grade problems than other models. Overall speaking, VPGTrans and MiniGPT4-13B are the superior multimodal models from the evaluation results.

\begin{table*}  
\centering  
\caption{Summary of multi-modal GPT models.}  

\begin{tabular}{l|l|l|l}  
\toprule
Method & \#Param &LLM &Vision Backbone \\
\midrule
MiniGPT4-7B~\cite{zhu2023minigpt}   &7.8B&Vicuna-7B&ViT-G(EVA-CLIP) \\
MiniGPT4-13B~\cite{zhu2023minigpt}  &14.1B&Vicuna-13B&ViT-G(EVA-CLIP) \\
LLaVA-7B~\cite{liu2023visual}       &6.7B&LLaMA-7B&ViT-L(CLIP) \\
LLaVA-13B~\cite{liu2023visual}      &13.0B&LLaMA-13B&ViT-L(CLIP) \\
VPGTrans~\cite{zhang2023transfer}   &7.8B&Vicuna-7B&ViT-G(EVA-CLIP) \\
mPLUG-Owl~\cite{ye2023mplug}        &7.2B&LLaMA-7B&ViT-L(CLIP) \\
Otter~\cite{li2023otter}            &8.2B&LLaMA-7B&ViT-L(CLIP) \\
MultiModal-GPT~\cite{gong2023multimodal}&8.4B&LLaMA-7B&ViT-L(CLIP) \\

\bottomrule
\end{tabular}  

\vspace{2mm}
\label{tab:7_mmllm_summary}  
\end{table*}

\begin{table*}[t]  
\centering  
\caption{Results (accuracy \%) on Science QA dataset.  Question classes: NAT = natural science, SOC = social science, LAN = language
science, TXT = text context, IMG = image context, NO = no context, G1-6 = grades 1-6, G7-12 = grades 7-12. \textcolor{orange!80!red}{Orange} indicates best performance, while \textcolor[HTML]{FEB8DD}{pink} indicates the second and third best.}  
\resizebox{!}{!}{
\begin{tabular}{l|ccc|ccc|cc}  
\toprule
\multirow{2}{*}{Method}    & \multicolumn{3}{c|}{Subject} & \multicolumn{3}{c|}{Context Modality} & \multicolumn{2}{c}{Grade} \\
     & NAT   & SOC   & LAN   & TXT   & IMG   & NO    & G1-6  & G7-12 \\ 
    \midrule
  \multicolumn{9}{l}{\it Representative \& SoTA methods with numbers reported in the literature } \\   
 Human~\cite{lu2022learn} &  90.23 & 84.97 & 87.48 & 89.60 & 87.50 & 88.10 & 91.59 & 82.42 \\
 GPT-3.5~\cite{lu2022learn}  &  74.64 & 69.74 & 76.00 & 74.44 & 67.28 & 77.42 & 76.80 & 68.89 \\
 GPT-3.5 w/ CoT~\cite{lu2022learn}   & 75.44 & 70.87 & 78.09 & 74.68 & 67.43 & 79.93 & 78.23 & 69.68 \\
GPT-4 & 84.06 & 73.45 & 87.36 & 81.87 & 70.75 & 90.73 & 84.69 & 79.10 \\
LLaMA-Adapter~\cite{zhang2023llama} & 84.37 &  88.30  &  84.36  & 83.72 &  80.32 &  86.90 &  85.83 &  84.05 \\
MM-CoT$_{Base}$~\cite{zhang2023multimodal} &  87.52 & 77.17 & 85.82 & 87.88 & 82.90 & 86.83 & 84.65 & 85.37 \\
MM-CoT$_{Large}$~\cite{zhang2023multimodal} & 95.91 & 82.00 & 90.82 & 95.26 & 88.80 & 92.89 & 92.44 & 90.31 \\
\midrule
\multicolumn{9}{l}{\it Fine-Tuning Results} \\

LLaVA~\cite{liu2023visual}      & 90.36 & 95.95 & 88.00 & 89.49 & 88.00 & 90.66 & 90.93 & 90.90 \\
LLaVA+GPT-4~\cite{liu2023visual} & 90.36 & 95.50 & 88.55 & 89.05 & 87.80 & 91.08 & 92.22 & 88.73 \\
LLaVA+GPT-4~\cite{liu2023visual} & 91.56 & 96.74 & 91.09 & 90.62 & 88.99 & 93.52 & 92.73 & 92.16 \\
     \midrule
 \multicolumn{9}{l}{\it 2-Shot Results} \\

Stanford Alpaca-7B & 41.47 & 27.67 & 47.18 & 41.35 & 34.01 & 45.37 & 39.24 & 41.53 \\
StableLM-Tuned-Alpha-7B &40.72	&42.05	&29.58	&47.00	&42.62	&37.04	&44.74	&40.53 \\
Databricks Dolly-v2-7b & 41.61 & 28.12 & 47.82 & 43.11 & 36.59 & 43.48 & 40.60 & 40.01 \\
Databricks Dolly-v2-12b & 43.34 & 31.05 & 49.55 & 44.53 & 38.08 & 45.99 & 42.55 & 42.06 \\
Alpaca-Lora &43.52	&32.40	&51.45	&42.86	&36.94	&50.31	&43.61	&42.58 \\
Vicuna-7B-v0 &41.79	&31.27	&49.09	&42.72	&37.58	&45.51	&41.74	&41.00 \\
Vicuna-13B-v0 &46.09	&34.42	&45.45	&45.70	&40.01	&46.34	&44.82	&41.07 \\
Vicuna-7B-v1 &41.61	&27.33	&45.09	&42.18	&35.70	&43.00	&39.35	&39.82 \\ 
Vicuna-13B-v1 &\cellcolor[HTML]{FEE8DD} 46.94	& \cellcolor[HTML]{FEE8DD} 44.09	&47.18	& \cellcolor[HTML]{FEE8DD} 47.21	& \cellcolor[HTML]{FEE8DD} 45.07	&46.69	&\cellcolor[HTML]{FEE8DD} 48.02	&43.51 \\
MultiModal-GPT~\cite{gong2023multimodal}& 41.43 & 26.43 & 48.36 & 41.98 & 35.35 & 44.88 & 40.38 & 39.55 \\
LLaVA-7B~\cite{liu2023visual}       & 43.87 & 27.45 & 48.18 & 44.53 & 35.20 & 46.83 & 41.52 & 41.60 \\
LLaVA-13B~\cite{liu2023visual}      & 44.80 & 31.38 & 51.73 & 43.70 & 35.35 & \cellcolor[HTML]{FEE8DD} 52.06 & 44.35 & 42.78 \\
VPGTrans~\cite{zhang2023transfer}   & \cellcolor[HTML]{FEE8DD} 50.84 & \cellcolor[HTML]{FEE8DD} 42.18 & 
\cellcolor[HTML]{FEE8DD} 53.73 & \cellcolor[HTML]{FEE8DD} 52.39 & \cellcolor[HTML]{FEE8DD} 47.40 & \cellcolor[HTML]{FEE8DD} 50.73 & \cellcolor[HTML]{FEE8DD} 50.26 & \cellcolor[HTML]{FEE8DD} 48.91 \\
MiniGPT4-7B~\cite{zhu2023minigpt}   & 45.91	& 34.98	& \cellcolor[HTML]{FEE8DD} 53.00	& 46.29	& 39.71	& 50.54	& 45.34	& \cellcolor[HTML]{FEE8DD} 45.68 \\
MiniGPT4-13B~\cite{zhu2023minigpt}  & \cellcolor[HTML]{FC8D59} 55.95	& \cellcolor[HTML]{FC8D59} 52.31	& \cellcolor[HTML]{FC8D59} 55.00	& \cellcolor[HTML]{FC8D59} 55.77	& \cellcolor[HTML]{FC8D59} 52.95	& \cellcolor[HTML]{FC8D59} 54.43	& \cellcolor[HTML]{FC8D59} 57.71	& \cellcolor[HTML]{FC8D59} 49.97 \\
\bottomrule
\end{tabular}  
}
\vspace{2mm}
\label{tab:7_sqa_benchmark}  
\end{table*}

\begin{table*}[t]  
\centering  
\caption{Results (accuracy \%) on Science QA dataset with image context.  Question classes: NAT = natural science, SOC = social science, LAN = language
science, G1-6 = grades 1-6, G7-12 = grades 7-12. \textcolor{orange!80!red}{Orange} indicates best performance, while \textcolor[HTML]{FEB8DD}{pink} indicates the second and third best.}  
\begin{tabular}{l|ccc|cc}  
\toprule
\multirow{2}{*}{Method}  & \multicolumn{3}{c|}{Subject} & \multicolumn{2}{c}{Grade} \\
                                    & NAT   & SOC   & LAN   & G1-6  & G7-12 \\ 
\midrule
Random Select                       & 40.45 & 25.39 & 40.91 & 33.80 & 37.07  \\ 
\midrule
MultiModal-GPT~\cite{gong2023multimodal}& 42.02 & 24.21 & \cellcolor[HTML]{FEE8DD} 45.45 & 36.60 & 32.31 \\
mPLUG-Owl~\cite{ye2023mplug}            & 45.41 & 27.23 & 36.36 & 40.17 & 33.84 \\
LLaVA-7B~\cite{liu2023visual}           & 43.84 & 25.52 & 38.64 & 38.21 & 33.33 \\
LLaVA-13B~\cite{liu2023visual}          & 42.51 & 34.69 & 43.18 & 40.38 & 37.59 \\
Otter~\cite{li2023otter}                & 43.34 & 32.46 & 34.09 & 39.75 & 37.24 \\
VPGTrans~\cite{zhang2023transfer}       & \cellcolor[HTML]{FEE8DD} 52.36 & \cellcolor[HTML]{FEE8DD} 39.14 & \cellcolor[HTML]{FC8D59}{56.82} & \cellcolor[HTML]{FEE8DD} 48.92 & \cellcolor[HTML]{FC8D59} 43.88 \\
MiniGPT4-7B~\cite{zhu2023minigpt}       & \cellcolor[HTML]{FEE8DD} 46.15 & \cellcolor[HTML]{FEE8DD} 35.08 & 38.64 &\cellcolor[HTML]{FEE8DD}  42.13 & \cellcolor[HTML]{FEE8DD} 40.99 \\
MiniGPT4-13B~\cite{zhu2023minigpt}      & \cellcolor[HTML]{FC8D59}{54.84} & \cellcolor[HTML]{FC8D59}{50.13} & \cellcolor[HTML]{FEE8DD} 50.00 & \cellcolor[HTML]{FC8D59}{57.10} & \cellcolor[HTML]{FEE8DD}{42.86} \\
\bottomrule
\end{tabular}  
\vspace{2mm}

\label{tab:7_sqa_img_benchmark}  
\end{table*}
\section{Scientific GPT Models and Evaluation}
\label{sec8:sci}

Scientific research spans various domains and necessitates powerful language models that can effectively understand and generate text in specialized fields. In recent years, the advent of large-scale language models that are initially developed for general language understanding and generation tasks have been adapted and finetuned to cater specifically to scientific research, such as drug discovery and material design. In this section, we delve into the realm of scientific GPT models and their evaluation.

\subsection{Model Summary}
In this part, we briefly introduce models specifically designed for scientific research. We focus on the following aspects:
\begin{enumerate}
    \item \textbf{Model Structure}: Among scientific models, Transformers \cite{DBLP:conf/nips/VaswaniSPUJGKP17} are also the universal architecture of the models. Usually, the models in the scientific domain favor decoder-only adaptation for generation and discovery while the encoder-based models are less. 
    \item \textbf{Pretraining Dataset}: For scientific GPTs, additional modalities, like drugs, proteins, and DNA sequences, are usually leveraged besides text. Taking Galactica \cite{DBLP:journals/corr/abs-2211-09085} as an example, the pretraining dataset is a mixture of text, code, Simplified
molecular-input line-entry system (SMILES) sequences \cite{DBLP:journals/jcisd/Weininger88}, amino acid sequence, DNA sequence, and other modalities. 
    \item \textbf{Pretraining Tasks}: Most of the scientific GPT models adopt the pretraining tasks the same as the general language models, such as next token prediction and masked language modeling. It should be noted that the pretraining tasks are not just for the text but for a wide variety of scientific modalities.
\end{enumerate}

Next, we present the details of several scientific GPT models, i.e., BioGPT \cite{DBLP:journals/bib/LuoSXQZPL22}, Galactica \cite{DBLP:journals/corr/abs-2211-09085}, BiomedGPT \cite{DBLP:journals/corr/abs-2305-17100}, MolXPT \cite{DBLP:conf/acl/LiuZXW0QZL23}, MolFM \cite{luo2023molfm} and MolT5 \cite{DBLP:conf/emnlp/EdwardsLRHCJ22}.

\textbf{BioGPT} \cite{DBLP:journals/bib/LuoSXQZPL22} is a domain-specific decoder-only Transformer language model pre-trained on biomedical literature. The pretraining objective is the next token prediction. It outperforms most of the models in the biomedical domain by excelling in both discriminative tasks and text generation. For example, it largely surpasses other models' performance in PubMedQA~\cite{DBLP:conf/emnlp/JinDLCL19}. 

\textbf{Galactica} \cite{DBLP:journals/corr/abs-2211-09085} is a state-of-the-art language model designed to address information overload in scientific research. It utilized a Transformer architecture in a decoder-only setup. It was trained on a wide variety of modalities, i.e., text, \LaTeX, code, SMILES, amino acid sequence, and DNA sequence.
Trained on a large corpus of scientific literature, reference material, and knowledge bases (106 billion tokens), Galactica excels in storing, combining, and reasoning about scientific knowledge. It outperforms most of the existing models in a range of scientific tasks, including technical knowledge probes and reasoning tasks.

\textbf{BiomedGPT} \cite{DBLP:journals/corr/abs-2305-17100} is an encoder-decoder multimodal molecular foundation model with 1.6B parameters. It can accommodate various modalities, such as CT images and clinical notes, etc. It contains three types of pretraining tasks: (1) Vision \& Language (Captioning, Visual Question Answering); (2) Vision (Detection, Image Filling); (3) Language (Masked Language Modeling).


\textbf{MolXPT} \cite{DBLP:conf/acl/LiuZXW0QZL23} is a decoder-only transformer-based model pretrained on heterogeneous data including scientific text, SMILES sequences, and “wrapped” sequences between SMILES and text. Its pre-training objective function is also the next token prediction.


\textbf{MolT5} \cite{DBLP:conf/emnlp/EdwardsLRHCJ22} is an encoder-decoder transformer model pretrained with unlabeled text and molecule strings. It leverages two pretraining tasks, i.e., molecule captioning and text-based de novo molecule generation.

\subsection{Benchmark Evaluations}

\subsubsection{Dataset Summary}

\begin{table*}[!ht]
\centering
\caption{Overview of datasets for large language models}
\label{tab:7_sci_dataset}
\begin{tabular}{lllcl}
\toprule
Corpora & Domain & Size & Latest updated time & Link \\
\midrule
MedQA \cite{DBLP:journals/corr/abs-2009-13081} & Biomedicine & $\sim$61k    &    2021.3       &   \url{https://github.com/jind11/MedQA}   \\
MedMCQA \cite{pal2022medmcqa} & Biomedicine & $\sim$194k & 2022.5 & \url{https://github.com/medmcqa/medmcqa}\\
PubMedQA \cite{DBLP:conf/emnlp/JinDLCL19} & 
 Biomedicine & $\sim$212k &  2021.3 & \url{https://pubmedqa.github.io}\\
NLPEC \cite{DBLP:conf/emnlp/LiHCPW20} & Biomedicine & $\sim$2.1k & 2021.3 & \url{http://112.74.48.115:8157/}\\
SciQ \cite{welbl2017crowdsourcing} & Physics, Chemistry, and Biology & $\sim$14k &  2023.6 & \url{https://huggingface.co/datasets/sciq} \\

\bottomrule
\end{tabular}
\end{table*}

We evaluate the models on the following datasets in the biomedical and physics domains: MedQA \cite{DBLP:journals/corr/abs-2009-13081}, MedMCQA \cite{pal2022medmcqa}, PubMedQA \cite{DBLP:conf/emnlp/JinDLCL19}, NLPEC \cite{DBLP:conf/emnlp/LiHCPW20} and SciQ \cite{welbl2017crowdsourcing}. An overview of these datasets can be seen in Table \ref{tab:7_sci_dataset}.

\textbf{MedQA} \cite{DBLP:journals/corr/abs-2009-13081} is a dataset specifically designed for medical question-answering tasks. It covers a wide range of questions related to the medical field, encompassing various topics such as diseases, symptoms, treatments, and healthcare procedures.  The dataset undergoes careful curation to ensure relevance and accuracy, making it a valuable resource for the development and evaluation of machine learning models in the medical domain.

\textbf{MedMCQA} \cite{pal2022medmcqa} stands for Medical Multiple-Choice Question Answering. It focuses on the task of answering multiple-choice questions in the medical domain. The questions are designed to test the understanding of medical concepts and require selecting the most appropriate answer from a given set of options. MedMCQA provides a benchmark for evaluating the performance of machine learning models in handling medical multiple-choice question answering tasks.

\textbf{PubMedQA} \cite{DBLP:conf/emnlp/JinDLCL19} is a dataset specifically tailored for biomedical question answering. It is created using a large-scale corpus of scientific articles from PubMed, which is a widely used resource for biomedical research. The dataset comprises questions based on real-world information needs in the biomedical domain, and the answers are derived from the scientific literature. PubMedQA serves as a valuable resource for developing and evaluating question answering systems in the biomedical field.

\textbf{NLPEC} \cite{DBLP:conf/emnlp/LiHCPW20} is a dataset that utilizes the National Licensed Pharmacist Examination in China as a source of questions. Specifically, it focuses on the pharmacy comprehensive knowledge and skills section of the exam, which consists of 600 multiple-choice problems across four categories. The dataset includes examples from the past five years (2015-2019) of this section, serving as the test set while excluding questions of the multi-answer type.

\textbf{SciQ} \cite{welbl2017crowdsourcing} is a dataset comprising 13,679 crowdsourced science exam questions. These questions cover various subjects such as Physics, Chemistry, and Biology. Each question is presented in a multiple-choice format, with four answer options available. A majority of the questions include an accompanying paragraph that offers supporting evidence for the correct answer.

\begin{table*}[!htbp]
\centering
\rotatebox[origin=c]{90}{
\begin{varwidth}{\textheight}
\caption{Results (accuracy \%) on MedQA \cite{DBLP:journals/corr/abs-2009-13081} for different models and few shot settings. The asterisk (*) indicates exceeding the maximum token limit. "/" denotes language not supported.  "-" denotes data missing, due to the absence of relevant data in the literature. \textcolor{orange!80!red}{Orange} indicates best performance, while \textcolor[HTML]{FEB8DD}{pink} indicates the second and third best.}
\label{tab:medqa}
\tabcolsep=7pt
\begin{tabular}{l|cccc|cccc|cccc|cccc}
\toprule
\multirow{2}{*}{Model} & \multicolumn{4}{c}{Mainland China (5-option)} & \multicolumn{4}{c}{Taiwan (4-option)}        & \multicolumn{4}{c}{US (5-option)} & \multicolumn{4}{c}{US (4-option)} \\
                       & 0 shot  & 1 shot & 3 shot & 5 shot & 0 shot & 1 shot & 3 shot & 5 shot & 0 shot & 1 shot & 3 shot & 5 shot & 0 shot & 1 shot & 3 shot & 5 shot \\
\midrule
GPT-4 & 71.07 & - & - & 75.31 & 82.17& - & -& 84.57 & 74.71&-&-&78.64 & 78.87&-&-&81.38 \\
GPT-3.5 & 40.31 & - & - & 44.89 & 50.60 & - & - & 53.72 & 44.62 & -&-& 47.05 & 50.82 & - & - & 50.82\\
\midrule
\multicolumn{17}{l}{\it Results with standard prompts} \\
LLaMA-7B \cite{touvron2023llama} & 22.12 & 22.30 & 21.89 & 22.07 & 26.47 & 26.89 & 26.82 & 27.18 & 20.19 & 21.37 & 21.05 & 21.52 & 27.42 & 27.34 & 27.18 & 27.97 \\
LLaMA-13B \cite{touvron2023llama} & 17.83 & 21.95 & 22.74 & 22.53 & 23.71 & 24.42 & 24.42 & 24.42 & 27.26 & 25.69 & 26.47 & 26.08 & 30.95 & 32.29 & 34.17 & 32.84 \\
LLama 2-7B \cite{touvron2023llama_2} & 21.51 & 21.40 & 21.40 & 21.83 & 26.04 & 27.11 & 25.69 & 25.76 & 25.06 & 23.88 & 24.43 & 23.17 & 31.66 & 32.21 & 37.31 & 37.94 \\
LLama 2-13B \cite{touvron2023llama_2} & 27.23 & 25.63 & 24.40 & 24.64 & \cellcolor[HTML]{FEE8DD} 30.36 & 29.79 & \cellcolor[HTML]{FEE8DD} 30.29 & 29.79 & \cellcolor[HTML]{FEE8DD} 33.15 & 30.64 & 30.16 & 28.91 & 37.39 & \cellcolor[HTML]{FEE8DD} 41.32 & \cellcolor[HTML]{FC8D59}{42.42} & \cellcolor[HTML]{FEE8DD} 42.18 \\
ChatGLM-6B \cite{du2022glm, zeng2023glm-130b}             & \cellcolor[HTML]{FEE8DD}33.74   & \cellcolor[HTML]{FEE8DD}33.74  & 31.70  & 31.49  & 28.80  & 29.93  & 29.16  & 29.16  & 25.22  & 25.22  & 25.06  & 25.69  & 29.46  & 28.59  & 29.69  & 29.69  \\
Stanford Alpaca-7B \cite{alpaca}    &    22.80     &    24.69    &    24.58    &   24.34     &    27.81    &    29.30    &    28.59    &    29.58    &   25.77     &    25.69    &   28.12   &    27.73    &    31.97    &    30.56    &    31.58    &    33.31    \\
Alpaca-LoRA \cite{alpaca-lora} & 20.34 & 21.34 & 20.78 & 21.19 & 24.27 & 25.62 & 25.62 & 25.12 & 24.59 & 23.49 & 23.64 & 24.82 & 27.02 & 28.2 & 26.0 & 27.81 \\
Vicuna(FastChat)-7B \cite{vicuna} & 24.69 & 25.13 & 25.69 & 25.8 & 28.8 & 30.15 & 28.95 & 29.23 & 24.35 & 24.35 & 25.37 & 24.59 & 32.21 & 34.41 & 32.76 & 34.72 \\
Vicuna(FastChat)-13B \cite{vicuna} & 23.18 & 26.88 & 25.98 & 26.53 & 30.01 & 28.95 & 29.51 & 28.73 & 28.04 & 31.97 & 32.44 & 32.05 & 32.52 & 36.21 & 35.51 & 35.27 \\
StableLM-Tuned-Alpha-7B \cite{stablelm} & 18.8 & 19.21 & 19.67 & 19.82 & 26.4 & 26.89 & 26.47 & 26.26 & 20.97 & 20.58 & 21.68 & 21.37 & 24.74 & 23.72 & 24.19 & 24.43 \\
Databricks Dolly-v2-7b \cite{dolly} & 18.27 & 17.98 & 18.39 & 18.07 & 23.64 & 24.98 & 24.2 & 23.78 & 21.6 & 19.8 & 20.66 & 21.6 & 24.98 & 25.22 & 24.43 & 24.67 \\
Databricks Dolly-v2-12b \cite{dolly} & 19.0 & 20.11 & 20.2 & 20.52 & 21.8 & 25.62 & 26.47 & 24.98 & 17.28 & 20.82 & 20.66 & 20.66 & 24.19 & 26.08 & 26.32 & 25.06 \\
MOSS \cite{moss} & 25.13 & 25.6 & 26.56 & 23.96 & 27.81 & 28.38 & 28.52 & 29.16 & 21.45 & 22.47 & 21.05 & 21.76 & 27.57 & 26.94 & 26.32 & 25.92 \\
Open-Assistant \cite{openassistant} & 20.9 & 20.49 & 19.96 & 19.91 & 26.89 & 27.53 & 27.25 & 27.67 & 18.46 & 20.03 & 18.77 & 19.01 & 23.1 & 23.33 & 23.1 & 23.17 \\
Openchatkit \cite{openchatkit} & / & / & / & / & / & / & / & / & 23.64 & 19.64 & 19.8 & 20.11 & 26.94 & 27.42 & 27.57 & 27.42 \\
BELLE-7B \cite{belle} & \cellcolor[HTML]{FC8D59}{34.30} & 30.41 & 30.47 & 28.49 & \cellcolor[HTML]{FC8D59}{30.43} & 28.52 & 28.31 & 28.24 & 22.39 & 22.07 & 20.58 & 21.68 & 26.87 & 27.34 & 26.0 & 25.14 \\
MPT-7B-Instruct \cite{mpt7b} & / & / & / & / & /  & / & / & / & 19.4 & 20.35 & 20.27 & 21.76 & 23.72 & 26.16 & 26.16 & 25.84 \\
PandaLM-7B \cite{PandaLM} & 22.04 & 22.68 & 21.6 & 21.54 & 25.97 & 26.54 & 26.33 & 25.69 & 21.76 & 20.82 & 19.64 & 19.64 & 24.12 & 25.14 & 22.47 & 26.0 \\
RWKV(Pile)-7B \cite{peng2023rwkv} & 18.21 & 19.24 & 19.70 & 19.73 & 25.76 & 25.97 & 26.40 & 26.47 & 20.35 & 20.74 & 19.56 & 20.19 & 27.89 & 27.97 & 26.94 & 27.02 \\
h2oGPT-6.9B    \cite{h2ogpt}           & / & / & / & / & / & / & / & / & 19.32 & 19.48 & 19.40 & 19.87 & 27.97 & 28.12 & 26.71 & 28.04 \\
h2oGPT-12B  \cite{h2ogpt}      & / & / & / & / & / & / & / & / & 17.75 & 16.50 & 18.62 & 18.62 & 26.87 & 27.18 & 27.42 & 27.34 \\
RedPajama(Base)-7B \cite{RedPajama-INCITE}      & / & / & / & / & / & / & / & / & 21.60 & 20.82 & 20.58 & 20.27 & 26.24 & 26.94 & 25.06 & 23.88 \\
RedPajama(Instruct)-7B \cite{RedPajama-INCITE}  & / & / & / & / & / & / & / & / & 21.60 & 21.13 & 20.27 & 19.32 & 26.24 & 26.94 & 27.34 & 27.97 \\

Galactica-6.7B \cite{DBLP:journals/corr/abs-2211-09085} & / & / & / & / & / & / & / & / & 27.65 & 24.98 & 24.19 & 22.94 & 33.39 & 32.84 & 32.29 & 32.52 \\
\midrule
\multicolumn{17}{l}{\it Results with specific system meta instructions} \\
ChatGLM-6B \cite{du2022glm, zeng2023glm-130b} & 29.92 & 30.15 & 29.77 & 29.31 & 26.68 & 24.98 & 26.75 & 25.69 & 25.37 & 22.47 & 23.41 & 22.31 & 27.89 & 25.84 & 26.71 & 25.14 \\
Stanford Alpaca-7B \cite{alpaca}  & 18.27 & 17.98 & 17.57 & 18.18 & 25.83 & 26.54 & 25.83 & 25.76 & 24.9 & 23.88 & 24.51 & 26.16 & 31.97 & 29.46 & 29.69 & 32.05 \\
Alpaca-LoRA \cite{alpaca-lora} & 18.42 & 17.66 & 18.16 & 18.74 & 23.71 & 24.84 & 26.11 & 26.61 & 23.49 & 23.72 & 23.96 & 23.72 & 27.26 & 29.3 & 28.44 & 29.3 \\
Vicuna(FastChat)-7B \cite{vicuna} & 22.65 & 22.59 & 22.53 & 22.01 & 27.25 & 28.1 & 27.03 & 25.48 & 23.72 & 23.64 & 26.94 & 27.97 & 27.97 & 31.58 & 29.77 & 31.42 \\
Vicuna(FastChat)-13B \cite{vicuna}  & 21.37 & 20.93 & 22.65 & 23.61 & 27.74 & 27.03 & 27.95 & 28.52 & 27.97 & 30.87 &\cellcolor[HTML]{FEE8DD} 33.23 & \cellcolor[HTML]{FC8D59}{33.54} & 32.68 & 36.29 & 36.76 & 35.43 \\
StableLM-Tuned-Alpha-7B \cite{stablelm} & 17.63 & 17.57 & 17.51 & 17.57 & 25.41 & 23.85 & 23.78 & 23.85 & 21.05 & 21.52 & 21.92 & 21.68 & 27.18 & 23.1 & 23.96 & 24.35 \\
Databricks Dolly-v2-7b \cite{dolly} & 17.57 & 17.78 & 18.24 & 18.01 & 23.57 & 23.85 & 23.99 & 23.57 & 21.92 & 21.13 & 20.66 & 21.68 & 27.65 & 26.32 & 26.08 & 22.86 \\
Databricks Dolly-v2-12b \cite{dolly} & 18.45 & 17.57 & 18.39 & 18.3 & 23.92 & 23.5 & 23.85 & 23.99 & 18.46 & 19.8 & 20.42 & 18.85 & 25.84 & 23.57 & 27.34 & 24.27 \\
MOSS \cite{moss} & 24.99 & 25.57 & 26.8 & 26.53 & 26.75 & * & * & * & 23.17 & 23.25 & 23.33 & 23.64 & 28.12 & 27.49 & 27.81 & 27.02 \\
Open-Assistant \cite{openassistant} & 17.63 & 17.54 & 17.54 & 17.57 & 23.57 & 23.99 & 23.92 & 23.78 & 19.48 & 21.52 & 21.37 & 21.45 & 26.87 & 27.65 & 27.65 & 27.73 \\
Openchatkit \cite{openchatkit} & 21.22 & 20.55 & 19.64 & 19.85 & 26.33 & 26.68 & 26.68 & 26.75 & 21.45 & 19.72 & 20.5 & 19.72 & 22.15 & 20.9 & 23.49 & 23.57 \\
BELLE-7B \cite{belle} & 27.73 & 23.06 & 24.14 & 23.5 & 29.65 & 25.69 & 26.89 & 25.83 & 17.05 & 16.89 & 16.73 & 18.46 & 27.34 & 25.92 & 21.92 & 21.13 \\
MPT-7B-Instruct \cite{mpt7b} & 19.53 & 22.01 & 18.45 & 18.27 & 25.34 & 25.62 & 23.5 & 24.42 & 22.15 & 21.92 & 21.13 & 21.68 & 27.97 & 25.61 & 26.71 & 27.02 \\
   \bottomrule
\end{tabular}
\end{varwidth}
}
\end{table*}

\begin{table*}[!htbp]
\centering
\rotatebox[origin=c]{90}{
\begin{varwidth}{\textheight}
\caption{Results (accuracy \%) on MedMCQA \cite{pal2022medmcqa}, PubMedQA \cite{DBLP:conf/emnlp/JinDLCL19}, NLPEC \cite{DBLP:conf/emnlp/LiHCPW20} and Sciq \cite{welbl2017crowdsourcing} for different models and few shot settings. The asterisk (*) indicates exceeding the maximum token limit. "/" denotes language not supported.  "-" denotes data missing, due to the absence of relevant data in the literature. \textcolor{orange!80!red}{Orange} indicates best performance, while \textcolor[HTML]{FEB8DD}{pink} indicates the second and third best.}
\label{tab:medmcqa_pubmedqa_nlpec}
\tabcolsep=4pt
\begin{tabular}{l|cccc|cccc|cccc|cccc|cccc}
\toprule
\multirow{2}{*}{Model} & \multicolumn{4}{c}{MedMCQA} & \multicolumn{4}{c}{PubMedQA}        & \multicolumn{4}{c}{NLPEC (English)} & \multicolumn{4}{c}{NLPEC (Chinese)} & \multicolumn{4}{c}{SciQ}\\
                       & 0 shot  & 1 shot & 3 shot & 5 shot & 0 shot & 1 shot & 3 shot & 5 shot & 0 shot & 1 shot & 3 shot & 5 shot & 0 shot & 1 shot & 3 shot & 5 shot & 0 shot & 1 shot & 3 shot & 5 shot \\
\midrule
GPT-4 & 69.52 & - & - & 72.36 & 75.2 & - & - & 74.4 & - & - & - & - & - & - & - & - & - & - & - & -\\
GPT-3.5 & 50.08 & - & - & 51.02 & 71.6 & - & - & 60.2 & - & - & - & - & - & - & - & - & - & - & - & -\\
\midrule
\multicolumn{17}{l}{\it Results with standard prompts} \\
LLaMA-7B \cite{touvron2023llama} & 23.45 & 25.77 & 24.86 & 24.84 & 69.8 & 72.4 & 67.0 & 71.8 & 20.18 & 26.00 & 26.00 & 25.64 & 21.64 & 15.82 & 17.27 & 20.36 & 45.7 & 47.1 & 49.4 & 50.2 \\
LLaMA-13B \cite{touvron2023llama} & 35.52 & 34.78 & 35.72 & 35.67 & 59.0 & \cellcolor[HTML]{FEE8DD} 73.6 & 34.2 & 65.6 & 22.73 & 25.27 & 23.09 & 24.18 & 20.55 & 21.82 & 23.64 & 20.36 & 83.7 & 79.2 & 85.5 & 87.4 \\
LLama 2-7B \cite{touvron2023llama_2} & 33.56 & 32.27 & 33.13 & 30.03 & 57.2 & 70.2 & 33.2 & 48.8 & 20.55 & 21.45 & 23.27 & 25.82 & 20.55 & 20.00 & 17.64 & 20.91 & 76.1 & 86.8 & 87.9 & 87.2 \\
LLama 2-13B \cite{touvron2023llama_2} & 37.25 & 37.17 & \cellcolor[HTML]{FEE8DD} 38.90 & \cellcolor[HTML]{FC8D59}{38.94} & 55.6 & 72.2 & 70.0 & 68.2 & 25.09 & \cellcolor[HTML]{FEE8DD} 35.82 & \cellcolor[HTML]{FC8D59}{40.73} & \cellcolor[HTML]{FEE8DD} 38.36 & 24.55 & 27.09 & 30.91 & 31.64 & 92.5 & 94.2 & \cellcolor[HTML]{FC8D59}{94.5} & \cellcolor[HTML]{FEE8DD} 94.4 \\
ChatGLM-6B \cite{du2022glm, zeng2023glm-130b} & 32.44 & 30.77 & 30.48 & 30.72 & 63.4 & 65.4 & 67.4 & 68.0 & 31.27 & 30.0 & 29.64 & 29.82 & \cellcolor[HTML]{FEE8DD} 31.82 & 30.91 & \cellcolor[HTML]{FC8D59}{34.36} & 31.45 & 94.2 & 93.0 & \cellcolor[HTML]{FEE8DD}93.7 & \cellcolor[HTML]{FEE8DD}93.7\\
Stanford Alpaca-7B \cite{alpaca} & 28.47 & 28.5 & 27.23 & 27.99 & \cellcolor[HTML]{FC8D59}{74.0} & 68.8 & 71.8 & 72.8 & 29.45 & 28.0 & 25.82 & 22.91 & 20.18 & 18.91 & 17.45 & 20.0 & 80.9 & 64.7 & 66.0 & 74.7\\
Alpaca-LoRA \cite{alpaca-lora} & 24.7 & 26.34 & 26.34 & 26.3 & \cellcolor[HTML]{FEE8DD} 73.6 & 71.6 & 71.0 & 70.2 & 21.09 & 19.82 & 24.55 & 25.82 & 20.18 & 18.36 & 19.64 & 20.18 & 65.2 & 41.2 & 43.5 & 48.5\\
Vicuna(FastChat)-7B \cite{vicuna} & 32.03 & 33.8 & 32.75 & 31.91 & 59.8 & 55.8 & 69.4 & 70.6 & 23.82 & 30.36 & 30.0 & 28.55 & 20.91 & 23.64 & 18.55 & 23.27 & 89.3 & 87.3 & 88.4 & 87.8\\
Vicuna(FastChat)-13B \cite{vicuna} & 33.92 & 36.58 & 37.49 & 35.98 & 60.6 & 67.6 & 71.0 & 71.4 & 28.0 & 31.64 & 29.09 & 29.09 & 20.18 & 24.55 & 21.45 & 21.64 & 90.2 & 89.0 & 89.4 & 89.2\\
StableLM-Tuned-Alpha-7B \cite{stablelm} & 24.26 & 25.05 & 25.72 & 25.65 & 55.0 & 58.4 & 55.6 & 57.2 & 20.73 & 23.82 & 20.55 & 22.36 & 20.18 & 16.36 & 19.09 & 24.18 & 26.6 & 26.9 & 25.6 & 25.6\\
Databricks Dolly-v2-7b \cite{dolly} & 26.42 & 26.61 & 26.3 & 26.23 & 58.2 & 58.2 & 63.0 & 67.4 & 19.45 & 20.36 & 26.36 & 23.45 & 21.45 & 20.73 & 20.36 & * & 26.0 & 28.7 & 29.1 & 28.6\\
Databricks Dolly-v2-12b \cite{dolly} & 23.14 & 25.46 & 24.12 & 23.43 & 66.6 & 65.4 & 64.4 & 67.2 & 20.91 & 25.64 & 22.73 & 22.00 & 20.00 & 20.36 & 21.45 & * & 30.4 & 32.1 & 30.7 & 34.0\\
MOSS \cite{moss} & 31.51 & 31.2 & 27.25 & 26.11 & 39.0 & 38.2 & 49.2 & 35.0 & 27.27 & 26.00 & 23.45 & * & 24.91 & 26.73 & 24.91 & * & 82.5 & 80.6 & 80.9 & 75.9\\
Open-Assistant \cite{openassistant} & 27.35 & 25.27 & 24.65 & 24.53 & 54.4 & 56.6 & 56.6 & 58.0 & 19.64 & 20.55 & 21.09 & 18.91 & 20.55 & 14.91 & 18.00 & * & 24.0 & 25.7 & 26.0 & 25.5\\
Openchatkit \cite{openchatkit} & 28.71 & 29.45 & 30.89 & 31.99 & 59.4 & 59.2 & 20.0 & 33.0 & 19.82 & 18.55 & 18.55 & * & / & / & / & / & 28.5 & 28.3 & 24.8 & 24.2\\
BELLE-7B \cite{belle} & 29.38 & 25.82 & 25.99 & 26.32 & 61.4 & 42.2 & 42.0 & 39.4 & 30.73 & 28.73 & 22.18 & 21.82 & 28.55 & 32.55 & 28.18 & 27.82 & 92.8 & 83.0 & 80.7 & 81.7\\
MPT-7B-Instruct \cite{mpt7b} & 26.87 & 27.11 & 27.16 & 26.37 & 70.8 & 70.2 & 72.4 & 72.8 & 20.91 & 24.18 & 23.64 & * & / & / & / & / & 54.7 & 51.9 & 55.7 & 50.0\\
PandaLM-7B \cite{PandaLM} & 21.9 & 23.28 & 23.43 & 23.48 & 64.4 & 62.6 & 66.4 & 71.6 & 24.18 & 24.73 & 24.00 & 24.73 & 25.09 & 22.36 & 25.09 & 24.18 & 56.6 & 34.4 & 35.9 & 38.2\\

RWKV(Pile)-7B \cite{peng2023rwkv}           & 23.31 & 24.07 & 28.26 & 27.47 & 59.2 & 59.0 & 61.0 & 61.8 & 20.00 & 17.27 & 18.91 & 21.45 & 20.18 & 19.27 & 20.00 & 20.18 & 26.3 & 25.1 & 25.1 & 25.4\\
h2oGPT-6.9B \cite{h2ogpt}             & 26.94 & 27.73 & 28.52 & 30.15 & 58.0 & 55.4 & 52.4 & 56.6 & 19.82 & 17.45 & 22.91 & * & / & / & / & / & 26.3 & 25.4 & 24.2 & 23.5\\
h2oGPT-12B \cite{h2ogpt}              & 25.01 & 24.81 & 24.36 & 24.43 & 56.6 & 59.6 & 52.6 & 64.0 & 20.18 & 22.36 & 23.27 & * & / & / & / & / & 28.1 & 30.8 & 27.1 & 27.0\\
RedPajama(Base)-7B \cite{RedPajama-INCITE}      & 24.31 & 25.84 & 26.42 & 24.41 & 66.2 & 68.2 & 68.8 & 73.0 & 17.27 & 16.91 & 21.82 & * & / & / & / & / & 32.4 & 37.3 & 30.6 & 31.4\\
RedPajama(Instruct)-7B \cite{RedPajama-INCITE}  & 26.99 & 30.05 & 26.46 & 27.01 & 65.2 & 70.4 & 70.4 & 72.2 & 23.27 & 25.09 & 28.55 & * & / & / & / & / & 77.1 & 85.1 & 86.6 & 86.7\\
Galactica-6.7B \cite{DBLP:journals/corr/abs-2211-09085} & 26.68 & 27.71 & 26.87 & 25.96 & 60.8 & 63.6 & 63.6 & 64.0 & 20.36 & 23.09 & 21.82 & * & / & / & / & / & 77.0 & 79.6 & 82.6 & 82.2\\
\midrule
\multicolumn{17}{l}{\it Results with specific system meta instructions} \\
ChatGLM-6B \cite{du2022glm, zeng2023glm-130b} & 27.4 & 28.23 & 25.94 & 27.01 & 62.4 & 61.8 & 65.0 & 64.8 & 31.27 & 23.82 & 22.36 & 21.64 & \cellcolor[HTML]{FEE8DD} 33.45 & 31.27 & 29.45 & 31.64 & 90.5 & 83.3 & 76.3 & 78.4 \\
Stanford Alpaca-7B \cite{alpaca} & 32.7 & 29.38 & 26.49 & 26.7 & 58.2 & 61.0 & 72.8 & 69.8 & 21.09 & 24.91 & 20.91 & 20.55 & 20.36 & 20.18 & 20.0 & 16.73 & 80.9 & 64.7 & 66.0 & 74.7 \\
Alpaca-LoRA \cite{alpaca-lora} & 28.54 & 29.14 & 26.85 & 26.34 & 71.2 & 72.4 & 72.0 & 72.6 & 19.82 & 19.64 & 20.18 & 20.18 & 20.36 & 20.18 & 20.18 & 20.91 & 65.2 & 41.0 & 43.7 & 48.6 \\
Vicuna(FastChat)-7B \cite{vicuna} & 30.55 & 31.2 & 31.82 & 31.27 & 64.6 & 63.8 & 64.4 & 62.4 & 20.91 & 21.27 & 20.18 & 20.18 & 20.18 & 20.18 & 20.36 & 20.91 & 89.3 & 87.3 & 88.4 & 87.8 \\
Vicuna(FastChat)-13B \cite{vicuna} & 34.07 & 36.31 & \cellcolor[HTML]{FEE8DD} 37.99 & 36.46 & 63.8 & 66.4 & 69.6 & 71.0 & 22.36 & 26.36 & 24.0 & 20.55 & 20.73 & 22.18 & 20.73 & 18.0 & 90.2 & 89.0 & 89.4 & 89.2 \\
StableLM-Tuned-Alpha-7B \cite{stablelm} & 31.15 & 25.87 & 23.5 & 24.91 & 51.2 & 42.4 & 39.4 & 41.8 & 20.18 & 20.18 & 20.18 & 20.18 & 20.18 & 20.18 & 20.18 & 20.18 & 26.6 & 26.1 & 25.7 & 25.3 \\
Databricks Dolly-v2-7b \cite{dolly} & 31.22 & 28.42 & 26.58 & 26.85 & 62.8 & 68.2 & 68.8 & 70.2 & 20.55 & 22.0 & 19.82 & 23.45 & 20.18 & 20.18 & 19.45 & * & 26.1 & 28.9 & 27.7 & 28.6 \\
Databricks Dolly-v2-12b \cite{dolly} & 24.77 & 23.12 & 22.59 & 23.0 & 62.4 & 60.2 & 54.6 & 68.0 & 20.36 & 19.82 & 20.0 & 19.27 & 20.18 & 21.27 & 20.00 & * & 30.3 & 30.6 & 30.4 & 34.4\\
MOSS \cite{moss} & 31.51 & 31.2 & 27.25 & 26.11 & 48.4 & 57.4 & 68.6 & 65.4 & 26.0 & 26.18 & 24.36 & * & 26.36 & 25.27 & 23.09 & * & 81.6 & 87.8 & 88.0 & 87.5\\
Open-Assistant \cite{openassistant} & 27.35 & 25.27 & 24.65 & 24.53 & 57.6 & 53.2 & 55.2 & 52.0 & 20.18 & 20.73 & 20.73 & 20.36 & 20.36 & 20.18 & 20.18 & * & 23.2 & 24.6 & 25.5 & 24.7\\
Openchatkit \cite{openchatkit} & 28.71 & 29.45 & 30.89 & 31.99 & 57.4 & 64.8 & 14.2 & 36.4 & 20.0 & 20.18 & 20.0 & * & / & / & / & / & 30.3 & 27.9 & 27.7 & 26.4\\
BELLE-7B \cite{belle} & 29.38 & 25.82 & 25.99 & 26.32 & 56.8 & 55.6 & 60.6 & 58.2 & 27.82 & 19.45 & 20.73 & 15.64 & 28.0 & 24.91 & 21.45 & 20.0 & 79.7 & 27.3 & 46.5 & 40.2\\
MPT-7B-Instruct \cite{mpt7b} & 26.87 & 27.11 & 27.16 & 26.37 & 55.6 & 55.2 & 55.6 & 57.0 & 18.55 & 16.18 & 16.73 & * & / & / & / & / & 50.3 & 24.1 & 23.8 & 23.7\\
\bottomrule
\end{tabular}
\end{varwidth}
}
\end{table*}

\subsubsection{Results}
Comprehensive analysis of the performance of different models on the MedQA, MedMCQA, PubMedQA, NLPEC, and SciQ datasets, considering various few-shot settings and regions, is presented in Table \ref{tab:medqa} and \ref{tab:medmcqa_pubmedqa_nlpec}. The results are reported in terms of accuracy percentages and categorized into two settings: "Results with standard prompts" and "Results with specific system meta instructions." In the ``Results with standard prompts" setting, the models were evaluated using standard prompts, and their accuracy percentages were recorded across different few-shot settings (0 shot, 1 shot, 3 shot, and 5 shot). In the "Results with specific system meta instructions" setting, the same models were evaluated, but with specific system meta instructions.

Comparing the results in the two settings, we first can observe that with specific system meta instructions, certain models, such as Stanford Alpaca-7B, Vicuna(FastChat)-13B, StableLM-Tuned-Alpha-7B, and Databricks Dolly-v2-7B, demonstrate better performance. These models exhibit enhanced capability when provided with specific system meta-instructions, indicating their sensitivity to such instructions and their ability to leverage them effectively for improved results. On the other hand, the opposite trend is observed in some other models, like BELLE-7B. These models do not benefit significantly from specific system meta instructions and may even experience a decrease in performance. This variation in model response highlights the importance of understanding the interaction between model architecture, training, and specific meta-instructions, as it can significantly impact performance outcomes. Further investigation into the underlying mechanisms and factors influencing these observations is warranted to gain deeper insights into the behavior of large language models in response to different instruction formats and prompts.

Table \ref{tab:medqa} presents performance results of various models on the MedQA dataset, focusing on different regions, namely Mainland China, Taiwan, and the US. In the Chinese version of the dataset, ChatGLM-6B and BELLE-7B outperform other models, achieving approximately 34\% and 30\% accuracy for "Mainland China (5-option)" and "Taiwan (4-option)" respectively. This suggests that these two models, specifically trained on Chinese corpus, exhibit better performance in Chinese question answering. For the United States version of the dataset, LLAMA 2-13B achieves superior performance, reaching approximately 33\% and 42\% accuracy for "US (5-option)" and "US (4-option)" respectively. These results highlight Galactica as the state-of-the-art scientific language model, surpassing most other models in the English version of the dataset.

In Table \ref{tab:medmcqa_pubmedqa_nlpec}, the same models were evaluated on the remaining datasets.  For the MedMCQA dataset, LLama 2-13B and Vicuna(FastChat)-13B outperform other models. Stanford Alpaca-7B and Alpaca-LoRA outperform the other models on the PubMedQA dataset. LLama 2-13B and ChatGLM-6B show superior performance on NLPEC (English/Chinese) and SciQ datasets.

Moreover, across different few-shot settings, there are instances of both increasing and decreasing results for the tested models. This indicates that (1) not all models will necessarily perform better when combined with few-shot settings, and (2) the performance does not necessarily increase as the number of few-shot examples increases.

Based on the experimental results presented in Table \ref{tab:medqa} and Table \ref{tab:medmcqa_pubmedqa_nlpec}, it is evident that models of such scales (6B, 7B, 13B) still struggle to handle the entire dataset, revealing a significant gap from reaching 100\% or even 50\% accuracy (except for PubMedQA). One of the limiting factors for these models is the number of parameters, which affects their capacity and generalization ability. More efforts are required to advance the field of AI-driven scientific research.

\section{Miscs}

\subsection{Chinese GPT Models}
Common GPT models are pretrained with mainly English corpus with a small portion of other languages such as Chinese, which may hinder their ability in Chinese tasks. Here we introduce some Chinese-enhanced GPT models.

\textbf{BELLE}~\cite{belle} is a Chinese-enhanced Chat model based on LLaMA. It first extends Chinese vocabulary and further pretrains LLaMA on 3.4 billion Chinese tokens. Then it uses self-instruct tuning~\cite{selfinstruct} on chat data generated by ~\cite{alpaca} and \cite{chatgpt}. The released model checkpoints include 7B.

\textbf{Chinese-LLaMA-Alpaca and Chinese-LLaMA-Alpaca-2
}~\cite{chinese-llama-alpaca} is a series of Chinese-enhanced base/Chat models based on LLaMA and LLaMA2. It enhances LLaMA's vocabulary by adding additional Chinese tokens to improve the model's encoding of Chinese semantics. Additionally, it further integrates pre-training using Chinese data and finetunes the model with Chinese instruction datasets. The released model checkpoints of Chinese-LLaMA-Alpaca include 7B, 13B, and 33B. The released model checkpoints of Chinese-LLaMA-Alpaca-2 include 7B.

\textbf{Chinese-Vicuna}~\cite{leng2023chinese-vicuna} is a Chinese Instruction-following model based on LLaMA. It aims to build methods for tuning Chinese LLaMA models to follow instructions. These methods can be trained on a single Nvidia RTX-2080TI GPU and can be used to create multi-round chatbots to be trained on a single Nvidia RTX-3090 GPU with a context length of 2048. The released model checkpoints include 7B and 13B.

\textbf{Linly-Chinese-LLaMA}~\cite{linly} is a Chinese-enhanced foundation model based on LLaMA utilizing Chinese and Chinese-English parallel incremental pre-training to transfer its powerful language abilities from English to Chinese. The release model checkpoints include 7B, 13B, and 33B.  Linly-ChatFlow~\cite{linly} is the self-instruct tuned chat model based on Linly-Chinese-LLaMA. The release model checkpoints include 7B and 13B.

\textbf{Panda}~\cite{jiao2023panda} is a series of Chinese-enhanced models further pretrained on about 15 million Chinese sequences based on LLaMA. The released model checkpoints include 7B and 13B. Panda-Instruct is a series of instruction-tuned models on six Chinese datasets based on Panda. The released model checkpoints include 7B and 13B.

\textbf{Luotuo}~\cite{luotuo} uses Chinese translated Alpaca~\cite{alpaca} 52k data and Guanaco dataset~\cite{guanaco} to conduct self-instruct tuning base on LLaMA. Besides, it chooses Lora~\cite{hu2021lora} tuning rather than full parameter finetuning. The released model checkpoints include 7B.

\textbf{Firefly}~\cite{Firefly} is a series of chat models tuned on 1.1M Chinese instruction samples across different tasks based on BLOOM~\cite{scao2022bloom}, Baichuai~\cite{baichuan}, and Ziya~\cite{fengshenbang}. The released model checkpoints include 1.4B, 2.6B, 7B, and 13B.

\textbf{DoctorGLM}~\cite{xiong2023doctorglm} is a healthcare-purpose LLM
based on ChatGLM-6B~\cite{du2022glm, zeng2023glm-130b}. The model is trained on databases of medical dialogues in Chinese with ChatGPT's help. The released model checkpoints include 6B.
 
\textbf{BenTsao}~\cite{wang2023huatuo} (also Huatuo) is an open-sourced Chinese biomedical LLM based on LLaMA. It is instruction-tuned on over 8k instruction data in the Chinese biomedical domain. The release model checkpoints include 7B.

\textbf{HuaTuoGPT} is an open-sourced Chinese medical LLM based on Baichuan~\cite{baichuan} and Ziya~\cite{fengshenbang}. It is supervised and finetuned on both distilled data from ChatGPT and real-world data from doctors. The released model checkpoints include 7B and 13B.

\textbf{Chinese Llama 2} is an open-sourced Chinese Chat model based on LLAMA 2~\cite{touvron2023llama_2}. It is finetuned on 10.1M instruction tuning samples.

\subsection{ChatGPT-like Models on Different Applications}
There are many NLP models powered by domain-specific applications. We list several applications in Table~\ref{tab:9_chatgptalternative}\footnote{Most GPT alternatives are initially collected by \url{https://writesonic.com/blog/chatgpt-alternatives/\#upcoming-chatgpt-alternativesnew-ai-tools}}.

\paratitle{Writing} here refers to generating content, including text-based but not limited to it. By providing prompts, these softwares encourage users to engage in creative writing and explore new ideas.

\paraitem{Chatsonic} by Writesonic is an NLP processor for conversational AI that specifically targets the limitations of ChatGPT. Powered by GPT-4, this system goes beyond text generation and possesses a range of capabilities. These include question answering similar to ChatGPT, real-time generation of trending factual context, creation of AI artwork, and understanding voice commands.

\paraitem{Jasper Chat}, an amiable AI chatbot, can help one generate ideas, improve one's content, and even bring a smile to one's face. It's specifically designed for business purposes such as marketing, sales, and other applications, distinguishing it from ChatGPT.

\paratitle{Search Engines} here refers to empowering search engines with large language models, just like the GPT-4 in Microsoft Bing. By doing this, LLMs have access to the web pages as external knowledge beyond parameters.

\paraitem{ChatSonic on Opera}. The search engine Opera is equipped with ChatSonic. With this integration, Opera can provide customized answers to one's specific queries by searching the web pages. 

\paraitem{NeevaAI} combines the capabilities of ChatGPT and other advanced language models with the Neeva search engine. By efficiently searching and analyzing a vast number of web pages, NeevaAI can produce a thorough response with relevant sources.

\paratitle{Coding} here refers to the ability to generate code. By providing prompts such as incomplete code or natural language comments, these softwares can generate code coherently can fulfill users' intent.

\paraitem{Copilot} leverages the OpenAI Codex~\cite{codex} to provide real-time suggestions of code snippets and complete functions from one's editor. Copilot has the ability to convert natural language prompts into coding suggestions that can be used in various programming languages.

\paraitem{Tabnine} is an AI assistant that enhances code delivery efficiency and ensures code security. It has the ability to automatically complete lines of code, propose complete function suggestions based on function declaration, and generate code blocks based on natural language comments.

\paraitem{Codewhisperer} is a code generation application developed by Amazon. It has the ability to generate code suggestions in real-time based on one's comments and existing code, annotate or filter code suggestions similar to open-source training data, and scan one's code to detect vulnerabilities.

\paratitle{Language Learning} here refers to the ability to help users learn languages such as grammar and tone correction.

\paraitem{Elsa} is developed to serve as an English Language Speech Assistant. This app is specifically designed to make improving one's English pronunciation an enjoyable and engaging experience.

\paraitem{DeepL Write} can correct grammar and punctuation errors, rephrase sentences, use precise wording, and choose the most appropriate tone for your text.

\paratitle{Research} here refers to the ability to help one in doing research, such as reading papers or providing domain-specific information.

\paraitem{Elicit} utilizes language models to streamline one's research processes, such as sections of literature review. It can identify pertinent articles even without an exact keyword match, condense the main findings of the paper related to one's query, and extract essential details from the articles.

\paraitem{Copilot in Azure Quantum} is an LLM tool powered by GPT-4, designed to aid quantum researchers in speeding up scientific exploration. With a user-friendly chat interface supplemented by relevant topic-specific data, it offers enhanced support for researchers.

\paraitem{ChatPDF} is an AI-powered tool that allows one to interact with PDF documents in a natural way. This makes it ideal for quickly extracting information or answering questions from large PDF files, such as manuals, essays, legal contracts, books, or research papers.

\paratitle{Productivity} here refers to the ability to improve one's daily work efficiency, such as taking notes or coping with Excel.

\paraitem{CoGram} offers the convenience of taking automatic notes during virtual conferences while prioritizing the privacy and security of one's data.

\paraitem{Otter} has the capability to generate meeting notes automatically by providing real-time transcription, recording audio, capturing slides automatically, and producing meeting summaries automatically.

\paraitem{Chatexcel} is an AI-powered tool that allows for interactive control of Excel through text chat. It is designed to simplify the workflow of accountants, teachers, and other professionals by allowing users to enter spreadsheet requirements and receive processed data with minimal additional steps.

\paraitem{AI Anywhere} serves as 's ultimate AI Copilot across all applications, not exclusively limited to websites, empowering one to effortlessly utilize ChatGPT without the need for copy-pasting.

\paratitle{Conversation} here refers to the ability to have human-like conversations with users.

\paraitem{Replika}, an AI-powered chatbot, acts as a companion and responds promptly to one's messages. It allows one to engage in conversations about life, romance, and other topics similar to what one would discuss with one's friends and family.

\paraitem{Character AI} similar to ChatGPT, enables users to engage in conversations with AI characters. With Character AI, one can have the opportunity to engage in conversations and interact with a wide range of personalities.


\paratitle{Building one’s own AI} refers to the ability to be one's own AI according to the user's own characteristics, rather than a one-for-all AI.

\paraitem{Botsonic AI chatbot} simplifies the process of consolidating all one's own data effortlessly, without requiring any coding. It can transform one's knowledge base into a chat-ready format by training ChatGPT using one's own data.

\begin{table*}  
\centering  
\caption{ChatGPT Alternatives on Different Applications}

\begin{tabular}{l|c|p{0.3 \columnwidth}|l}  
\toprule
Field & Software & Backbone & Url \\
\midrule
\multirow{2}*{Writing}
&ChatSonic &GPT-4 & \url{https://writesonic.com/chat} \\
&Jasper Chat &GPT 3.5 and others & \url{https://www.jasper.ai/chat} \\
\midrule
\multirow{2}*{Search Engines}
&ChatSonic on Opera&GPT-4 & \url{https://writesonic.com/chatsonic-opera} \\
&NeevaAI&ChatGPT & \url{https://neeva.com/} \\
\midrule
\multirow{3}*{Coding}
&Copilot &Codex & \url{https://github.com/features/copilot} \\
&Tabnine &GPT-2 & \url{https://www.tabnine.com/} \\
&Codewhisperer & - & \url{https://aws.amazon.com/cn/codewhisperer} \\
\midrule
\multirow{2}*{Languege Learning}
&Elsa & - & \url{https://elsaspeak.com/en} \\ 
&DeepL Write & - & \url{https://www.deepl.com/translator} \\
\midrule
\multirow{3}*{Research}
&Elicit & - & \url{https://elicit.org} \\
& ChatPDF  & ChatGPT  & \url{https://www.chatpdf.com/}
\\
&Copilot in Azure Quantum & GPT-4 & \url{https://quantum.microsoft.com/} \\ 
\midrule
\multirow{4}*{Productivity}
&CoGram & - & \url{https://www.cogram.com} \\ 
&Otter & - & \url{https://otter.ai} \\
&Chatexcel & - & \url{https://chatexcel.com/} \\
&AI Anywhere & ChatGPT, GPT-4 & \url{https://www.ai-anywhere.com/#/dashboard} \\
\midrule
\multirow{2}*{Conversation}
&Replika & A model with 774M parameters & \url{https://replika.com} \\
&Character AI & GPT-4 & \url{https://beta.character.ai} \\
\midrule
\multirow{1}*{Building one's own AI}
&Botsonic AI chatbot & GPT-4 & \url{https://writesonic.com/botsonic} \\

\bottomrule
\end{tabular}  

\vspace{2mm}
\label{tab:9_chatgptalternative}  
\end{table*}


\subsection{Tool Learning with Foundation Models}

Despite achieving remarkable successes, current LLM technologies still face certain urgent challenges in their journey toward developing an Artificial General Intelligence (AGI) system. These challenges can be discussed from the following perspectives: (1) Limited Scope: Although current LLMs have made significant advancements in NLP tasks, they are limited in their ability to process complex information from other modalities such as vision and speech. Their expertise lies primarily in text generation, and they lack the capability to handle more intricate forms of data. (2) Complex Task Handling: Real-world scenarios often involve complex tasks that consist of multiple sub-tasks. Current large language models are primarily trained based on existing data, so when generating text or solving tasks, they mainly rely on their own knowledge and experience, lacking the ability to leverage external resources to assist themselves.

Here we briefly introduce several works about tool learning with foundation models after the emergence of ChatGPT. For more related works and details of tool learning, we refer readers to a more comprehensive survey about tool learning~\cite{qin2023tool}. \textbf{Visual ChatGPT}~\cite{wu2023visual} is a system that gives ChatGPT the ability to understand and generate visual content. Instead of starting from scratch to train a new multi-modal ChatGPT, they have developed Visual ChatGPT by directly building upon ChatGPT and incorporating various Visual Foundation Models (VFMs). They introduce a Prompt Manager to convert diverse visual information into language format to communicate with ChatGPT. Similar works include \textbf{ViperGPT}~\cite{surís2023vipergpt} that prompts LLMs to produce Python code to answer complex visual queries and \textbf{MM-REACT}~\cite{yang2023mmreact} combines ChatGPT with a team of vision experts, enabling multimodal reasoning and action. \textbf{HuggingGPT}~\cite{shen2023hugginggpt} is a more general approach not limited to visual tasks. It utilizes the LLM as an interface to channel user requests toward expert models from Huggingface. \textbf{AutoGPT}\footnote{\url{https://github.com/Significant-Gravitas/Auto-GPT}.} is an extremely popular open-source application that can autonomously achieve whatever goal utilizing GPT-4.

\section{Conclusion and Future Directions}

In this survey, we conducted a comprehensive examination of alternative open-sourced models of large Generative Pre-trained Transformer (GPT) systems, with a specific focus on user-friendly and relatively small models that overcome the limitations posed by their larger counterparts. Our investigation highlighted the potential of these open-sourced GPT-like models to address the challenges associated with size, complexity, and closed development loops, while still maintaining high performance across various tasks and extending their capabilities to multimodal domains.

We present an analysis of benchmark evaluations in general question answering (QA) datasets, multimodal settings and scientific domains across zero-shot and few-shot settings. From the results, we can see that: (1) LLAMA2-7B, Vicuna-7B and Stanford Alpaca-7B outperform the other models in these benchmarks; (2) MiniGPT4-13B shows the best performance in multimodal QA dataset (i.e. ScienceQA); (3) LLAMA2-7B and Vicuna-13B show great superiority in scientific QA datasets (e.g., MedMCQA).

Furthermore, we explore the potential in human alignments by human evaluation. From the results, we can see that Vicuna-7B, ChatGLM and Moss show great superority, achieving higher Elo ratings. These evaluations also demonstrated that, despite their smaller size, part of these models, like vicuna-7B, can achieve commendable results in practical scenarios and are promising alternatives to their larger counterparts.

However, there are still some limitations of our survey. Next, we first present the limitations of this survey and then we discuss the challenges and future directions.

\subsection{Limitations}
While this survey paper provides an in-depth exploration of user-friendly and relatively small GPT-like models, it is important to acknowledge some limitations in our analysis. First and foremost, our focus was primarily on GPT models that excel in natural language processing (NLP) and extend to multimodal, scientific domains. However, we did not extensively cover the \textbf{mathematical capabilities} of GPT models or their potential for advanced mathematical reasoning and problem-solving. The mathematical prowess of GPT models could be a crucial area of research and development, with implications for scientific simulations, engineering, and various quantitative tasks.

Also, another aspect is the \textbf{safety} in GPT models. As AI models, especially the large language models, become increasingly powerful and ubiquitous, ensuring their responsible and ethical usage becomes paramount. Addressing concerns such as bias, fairness, unintended harmful, and unhealthy consequences in GPT models warrants dedicated research and discussion, but it falls beyond the scope of this particular survey. There are other works that delve into these ethical considerations and explore methods to make GPT models safer and more aligned with human values.

Additionally, for the \textbf{open-sourced efforts}, although we highlighted the ongoing open-sourced projects and initiatives for user-friendly GPT model reproduction and deployment, the rapidly evolving landscape of AI research means that newer models and developments may emerge beyond this survey. Therefore, it is essential for researchers and practitioners to continually stay updated with the latest advancements in the field.

One limitation is about the \textbf{evaluations} in this survey. Though we have put much effort into the benchmark evaluations and human evaluations of these open-sourced GPT models, this is far from enough. We mostly choose the question answering dadaists for the evaluation, while other datasets are left behind. Besides, the human evaluation may be a constraint to the questions we designed for these models, more larger and wider human evaluation is required for a thorough evaluation. 

Lastly, there are various different aspects to test the capabilities of the large language models, such as knowledge reasoning, planning ability, and tool manipulation. In this survey, we are not specific to all of these aspects since separate efforts on these capability evaluations should be made and there are some initiatives~\cite{chang2023survey}. Notably, there are also some large language model surveys, for example, a large and comprehensive survey from Renmin University of China~\cite{zhao2023survey}. Different from them, our focus is only on the open-sourced, user-friendly, and relatively smaller GPT models since they are the most convenient ones for individuals and school labs to continue researching and developing. We also conduct benchmark and human evaluations on these open-sourced models to provide some guidance for selecting the superior models in the future.

\subsection{Challenges and Future Directions}

While user-friendly and relatively small GPT models show promise, they also face several challenges that offer opportunities for future research and development. Here, we only provide some key points. 
\begin{itemize}
    \item One significant challenge is striking a delicate balance between model size and performance. As we aim for more accessible and efficient models, there exists a trade-off between reducing model complexity and preserving task performance. Exploring novel architectural designs, compression techniques, and knowledge distillation methods could pave the way for creating leaner yet powerful GPT alternatives.
    \item Another important area is tackling the issue of data efficiency. Large GPT models typically require vast amounts of data for effective training. However, data collection and annotation can be costly and time-consuming, making it challenging to deploy GPT models in resource-constrained settings. Developing techniques to enhance data efficiency, such as transfer learning across related tasks and domains, can significantly expand the practical applications of user-friendly GPT models.
    \item Interpretable and explainable AI is another vital aspect that demands attention. As GPT models grow in complexity, understanding their decision-making processes becomes increasingly challenging. Future research should focus on developing methods to provide human-readable explanations for model predictions and ensure transparency and trust in AI applications.
    \item The deployment of user-friendly GPT models in real-world scenarios requires addressing challenges related to privacy and security. Developing robust defenses against adversarial attacks and ensuring data privacy during model training and inference will be critical to fostering widespread adoption and public trust.
    \item To truly democratize AI, future work should aim to enhance the accessibility of user-friendly GPT models for non-experts and individuals from diverse backgrounds. Creating intuitive interfaces, documentation, and tools that allow users to interact with and benefit from GPT models without deep technical expertise will be essential in realizing the full potential of these models.
    \item One trend is to develop scientific foundation models. These models would be tailored specifically to cater to the needs of the scientific community. Scientific foundation models have the potential to revolutionize scientific research by providing AI-powered tools for data analysis, hypothesis generation, and simulation in various scientific disciplines. By incorporating domain-specific knowledge and scientific reasoning capabilities, these models can assist researchers in accelerating discoveries, gaining new insights, and solving complex scientific problems. The future development of user-friendly scientific foundation models will require close collaboration between AI experts and domain scientists, ensuring that the models not only exhibit high performance but also align with the scientific principles and standards of each discipline. Such models could pave the way for a new era of AI-driven scientific discoveries and contribute significantly to the advancement of general AI in scientific research and beyond.

\end{itemize}

In conclusion, the challenges identified above present exciting avenues for further research and development of user-friendly GPT models. By addressing these challenges, we can pave the way for a more inclusive and responsible AI ecosystem, where GPT models become valuable tools for a broader scientific community and society as a whole.

\section*{Acknowledgement}
We sincerely thank the following partners for helping with the human evaluations, Peiyan Hu, Feng Xu, Xuerui Su, and Bohan Wang.

\ifCLASSOPTIONcaptionsoff
  \newpage
\fi

\bibliographystyle{IEEEtran}
\bibliography{new_ref}

\end{document}